\documentclass[10pt,twocolumn,letterpaper]{article}

\usepackage{cvpr}              

\usepackage[dvipsnames]{xcolor}
\usepackage{graphicx}
\usepackage{amsmath}
\usepackage{amssymb}
\usepackage{booktabs}
\usepackage[table]{xcolor} 
\usepackage{capt-of}

\usepackage[dvipsnames]{xcolor}

\makeatletter
\renewcommand{\@maketitle}{
   \newpage
   \null
   \iftoggle{cvprrebuttal}{\vspace*{-.3in}}{\vskip .375in}
   \begin{center}
      \iftoggle{cvprrebuttal}{{\large \bf \@title \par}}{{\Large \bf \@title \par}}
      \iftoggle{cvprrebuttal}{\vspace*{-22pt}}{\vspace*{24pt}}{
        \large
        \lineskip .5em
        \begin{tabular}[t]{c}
          \iftoggle{cvprfinal}{
            \@author
          }{
            \iftoggle{cvprrebuttal}{}{
              Anonymous \confName~submission\\
              \vspace*{1pt}\\
              Paper ID \paperID
            }
          }
        \end{tabular}
        \par
      }
      \vskip .25em
      \ifx\@affiliation\@empty\else
        {\small \@affiliation \par}
      \fi
      \vspace*{6pt}
   \end{center}
}
\makeatother

\definecolor{cvprblue}{rgb}{0.21,0.49,0.74}
\usepackage[pagebackref,breaklinks,colorlinks,citecolor=cvprblue]{hyperref}

\def\paperID{234} 
\def\confName{3DV\xspace}
\def\confYear{2026\xspace}

\title{Frequency-Aware Gaussian Splatting Decomposition}

\author{Yishai Lavi$^{*}$\\
{\tt\small yishailavi@mail.tau.ac.il}
\and
Leo Segre$^{*}$\\
{\tt\small leosegre@mail.tau.ac.il}
\and
Shai Avidan\\
{\tt\small avidan@eng.tau.ac.il}
}

\hypersetup{urlcolor=cvprblue}

\affiliation{
\vspace{4pt}
\qquad \ 
Tel Aviv University\\
\qquad \
{\footnotesize\url{https://yishailavi.github.io/nerfstudio_lap/}}%
}

\begin{document}


\twocolumn[{%
\maketitle
\begin{center}
    \captionsetup{type=figure}
    \includegraphics[width=0.98\textwidth]{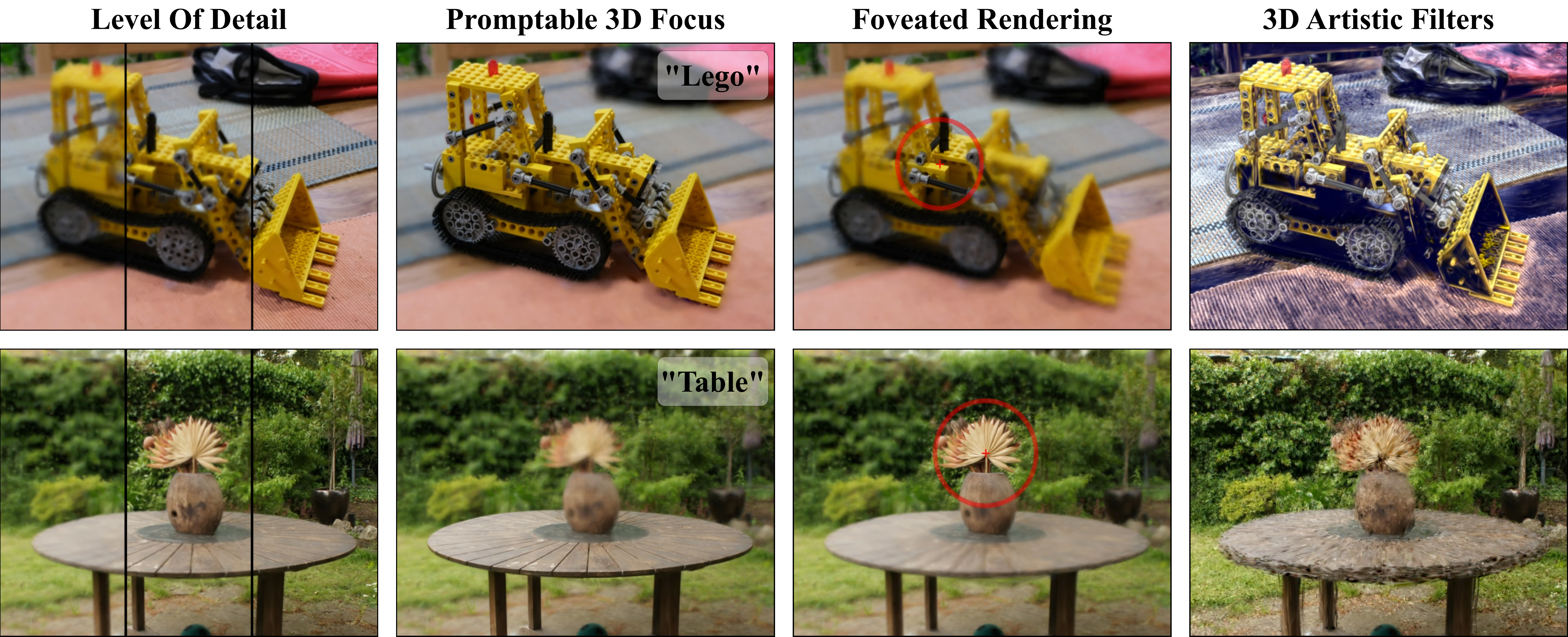}
    \captionof{figure}{\textbf{Frequency-Aware Decomposition for 3D Gaussian Splatting:}
    We introduce a frequency-aware decomposition for 3D Gaussian Splatting (3D-GS).
    The decomposition assigns different Gaussians to different levels of the Laplacian
    Pyramid of the input images. This enables Level of Detail (LOD) rendering,
    Promptable 3D focus, Foveated rendering, and 3D artistic filters.
    Our method achieves the highest PSNR, SSIM, and rendering speed among all
    LOD-capable baselines.
        \vspace{1em}
}
    \label{fig:teaser}
\end{center}
}]

\maketitle
\begingroup
\renewcommand\thefootnote{\fnsymbol{footnote}}
\footnotetext[1]{Equal contribution.}
\endgroup

\begin{abstract}
3D Gaussian Splatting (3D-GS) enables efficient novel view synthesis, but treats all frequencies uniformly, making it difficult to separate coarse structure from fine detail. Recent works have started to exploit frequency signals, but lack explicit frequency decomposition of the 3D representation itself. We propose a frequency-aware decomposition that organizes 3D Gaussians into groups corresponding to Laplacian-pyramid subbands of the input images. Each group is trained with spatial frequency regularization to confine it to its target frequency, while higher-frequency bands use signed residual colors to capture fine details that may be missed by lower-frequency reconstructions. A progressive coarse-to-fine training schedule stabilizes the decomposition.

Our method achieves state-of-the-art reconstruction quality and rendering speed among all LOD-capable methods.
In addition to improved interpretability, our method enables dynamic level-of-detail rendering, progressive streaming, foveated rendering, promptable 3D focus, and artistic filtering. 
Our code will be made publicly available.
\vspace{-0.5em}
\end{abstract}

\section{Introduction}

3D Gaussian Splatting (3D-GS) \cite{kerbl20233d} is an explicit, rasterization-based scene representation that delivers real-time, photorealistic view synthesis. But it lacks a notion of frequency: the millions of Gaussians are an unstructured pool, and beyond heuristics, there is no principled way to say which Gaussians carry low-frequency structure and which encode fine detail.
Recent works have begun to exploit frequencies for improved reconstruction quality - for example, FreGS \cite{zhang2024fregs} uses progressive frequency regularization to address over-reconstruction issues during densification. However, these approaches lack explicit frequency decomposition of the 3D representation itself, making it difficult to perform tasks such as level-of-detail rendering, efficient streaming, and control over different frequency components. Our goal is precisely to introduce this missing structural organization, which enables the above applications and more, as demonstrated in \Cref{fig:teaser} - 
\textbf{Level of Detail}: render the scene at a chosen level of detail for progressive rendering and streaming.
\textbf{Promptable 3D Focus}: keep selected objects sharp while blurring the rest based on a text prompt, in a 3D-consistent and interactive manner. 
\textbf{Foveated Rendering}: reduce detail in the peripheral area, increasing rendering speed. 
\textbf{3D Artistic Filters}: apply different artistic styles directly to the 3D scene.

We realize this structural organization by explicitly partitioning the Gaussians into a small number of \emph{groups} that correspond to specific subbands in the input images, leveraging the 2D frequency bands of the training images to define these groups. We implement this through a progressive, coarse-to-fine training scheme. First, we optimize a base level of Gaussians to explain the low-frequency content of the images;
we then introduce a new set of Gaussians for the next band and jointly optimize both levels, while constraining the earlier level to remain low pass using a frequency regularizer. This forces the newly added Gaussians to complement the reconstruction by capturing the missing higher-frequency residuals. Higher bands are added in the same manner. To model residuals faithfully, Gaussians in non-base levels use signed residual colors (RGB values can be positive or negative), allowing them to add or subtract color w.r.t. the accumulated rendering. The end result is a frequency-aware hierarchy in which each group renders a specific Laplacian-like subband, and any prefix of the hierarchy produces a valid, lower-detail reconstruction.

Using 2D image frequencies as a proxy is a deliberate design choice: it provides a stable and well-studied frequency basis, avoids the complexity of defining frequency for anisotropic splats that project differently across views, and, when enforced consistently across all views, yields a coherent segmentation of the underlying 3D content.

Our decomposition method achieves \textbf{state-of-the-art reconstruction metrics among level-of-detail and frequency-aware 3D-GS methods} on the standard benchmarks (\emph{Mip-NeRF360}, \emph{Tanks\&Temples}, and \emph{Deep Blending}) while preserving real-time rendering.

In short, we reframe level-of-detail not as mere hierarchical sparsification, but as \emph{frequency segmentation of the representation itself}. Our contributions are: (i) a frequency-aware decomposition of 3D-GS driven by image-space Laplacian bands, (ii) a progressive training strategy with signed residual colors and band-separating regularization, and (iii) comprehensive experiments demonstrating SOTA quality among LOD/frequency-aware methods together with diverse, frequency-driven applications.

\section{Related Work}

\begin{figure*}[h]
\begin{center}
    \includegraphics[width=\textwidth]{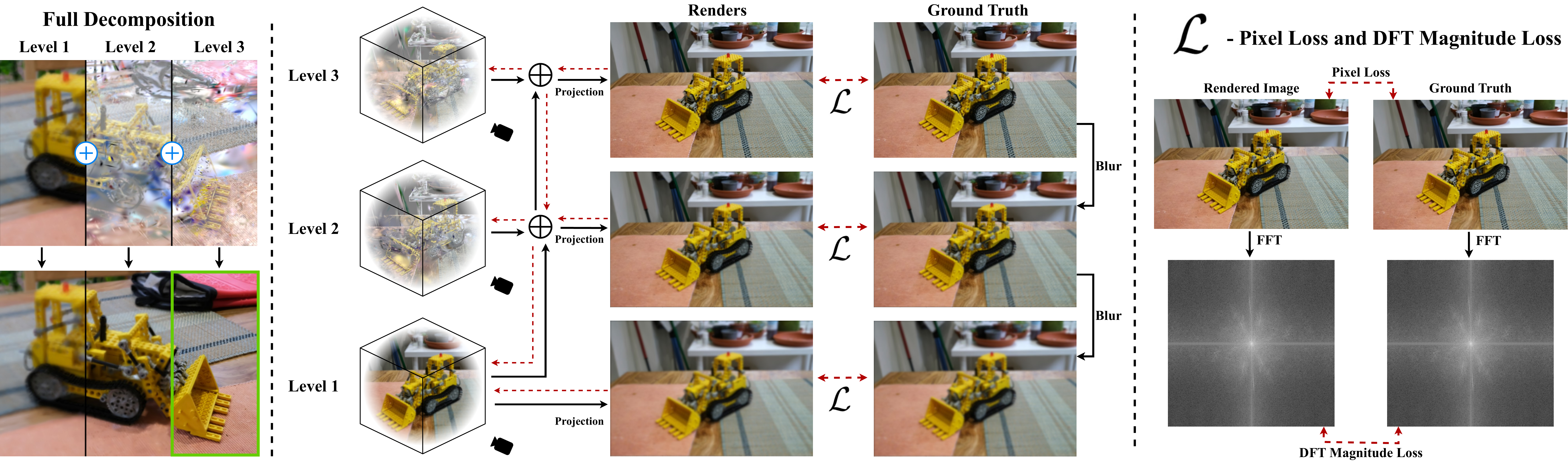} 
\end{center}
\caption{{\bf Frequency-Aware 3D Gaussian Splatting Overview:} We decompose 3D Gaussians into frequency-specific groups to construct a multi-scale representation. (Left) Each Gaussian group targets a specific frequency band: Level 1 captures low-frequency structure, while Levels 2-3 add mid and high-frequency details through signed residual colors. (Middle) During training, each accumulated level (L1, L1+L2, L1+L2+L3) is supervised against appropriately blurred ground truth images, with red dashed lines showing gradient flow that constrains lower levels to remain frequency-limited while higher levels capture residual details. (Right) Our dual loss combines standard pixel loss with DFT magnitude loss in frequency space, ensuring each level contributes meaningful frequency content and preventing frequency leakage between bands.}
\label{fig:short}
\end{figure*}

\subsection{Novel View Synthesis}
Novel view synthesis aims to generate photorealistic images from previously unseen viewpoints, typically given only sparse reference views. Early approaches primarily relied on explicit geometry or depth-based warping, such as layered depth images~\cite{shade1998layered}, view interpolation~\cite{chen1993view}, and light field modeling~\cite{levoy1996light}. Later methods incorporated learned priors and convolutional neural networks to predict novel viewpoints directly from input images~\cite{zhou2016view, park2017transformation}. A significant breakthrough came with Neural Radiance Fields (NeRF)~\cite{mildenhall2021nerf}, which represent scenes as continuous volumetric functions parameterized by multi-layer perceptrons, enabling high-quality view synthesis via volume rendering. NeRF has since inspired a series of high‑profile successors ~\cite{barron2021mip,muller2022instant,barron2023zip} that extend the original framework. However, the implicit MLP representation common to these models still imposes a non‑trivial computational load. Addressing efficiency from a different angle, 3D Gaussian Splatting (3D‑GS)~\cite{kerbl20233d} replaces the network with explicit oriented Gaussians and a fast differentiable rasterizer, markedly reducing overhead while preserving high‑quality rendering.

\subsection{Frequency-Aware Representations}
Many classical vision and graphics methods rely on frequency decompositions, such as Laplacian pyramids~\cite{burt1987laplacian}, discrete wavelet transforms~\cite{mallat1989theory}, and Fourier-domain filtering, enabling multiscale analysis. \cite{yang2022polynomial} introduce Polynomial Neural Fields (PNFs) and demonstrate frequency-based manipulation of neural fields using Fourier PNFs. Neural rendering methods integrate similar concepts using Fourier-feature embeddings~\cite{tancik2020fourier}, progressive-frequency training~\cite{zhu2023pyramid}, frequency-aware NeRFs for enhancing fine detail~\cite{zhang2025lookcloser} and wavelet-based NeRF architectures~\cite{xu2023wavenerf, khatib2024trinerflet}.

Within 3DGS, recent methods have started to exploit frequency signals. Mip‑Splatting~\cite{yu2024mip} enforces alias‑free rendering via analytic Gaussian filtering based on sampling rate limits. GES ~\cite{hamdi2024ges} additionally considers frequency aware loss using DoG. ~\cite{zeng2025frequency} links Gaussian scale and density to underlying frequency content, and ~\cite{10916987} extends this notion to dynamic scenes.
Similarly to our approach, FreGS \cite{zhang2024fregs} utilizes a Fourier-based loss, but applies it to guide Gaussian densification in a coarse-to-fine manner to mitigate over-reconstruction. In contrast, we use it to group Gaussians into distinct frequency subbands of the scene.

\subsection{Level of Detail in 3D Gaussian Splatting}  
Recent works have extended 3D Gaussian Splatting with level-of-detail mechanisms. Multi-Scale 3D-GS~\cite{yan2024multi} introduces multiple Gaussian scales to reduce aliasing artifacts low resolutions, while OCTree-GS~\cite{ren2024octree} organizes Gaussians within an octree structure~\cite{meagher1982geometric} to accelerate rendering across multiple scales and viewpoints.
Other methods, including ~\cite{kerbl2024hierarchical,liu2024citygaussian,liu2024citygaussianv2, wang2024pygs}, organize large-scale scenes using hierarchical data structures for efficient rendering, but focus primarily on scalability and massive scene handling rather than explicit frequency or detail separation.

LapisGS~\cite{shi2025lapisgslayeredprogressive3d} addresses adaptive streaming by training Gaussian layers sequentially on progressively higher-resolution inputs, allowing opacity changes between layers to enforce a coarse-to-fine structure. FLoD~\cite{seo2025flodintegratingflexiblelevel} organizes Gaussians into discrete levels according to their spatial extent, enabling customized resolution control through sequential optimization. Both approaches freeze most of the parameters of previously optimized levels before introducing new ones, relying on heuristic measures: resolution or spatial extent, to define their level-of-detail separation.

In contrast, our method explicitly defines Gaussian groups based on the frequency content of input images, jointly training all frequency groups progressively rather than sequentially freezing lower levels. We employ dedicated frequency-domain regularization to enforce clear spectral boundaries between these groups. Additionally, higher-frequency Gaussians use signed residual colors to encode fine details efficiently by subtracting colors from coarser frequency levels, similar to a Laplacian pyramid, thereby reducing the total number of Gaussians required.

\section{Method}

Our method is based on the foundational framework of 3D Gaussian splatting \cite{kerbl20233d}. By extending this method, we introduce a mechanism to group Gaussians based on their contribution to different frequency bands of the images. These groups, referred to as "levels", enable a hierarchical representation akin to the Image Laplacian Pyramid \cite{burt1987laplacian} in classical computer vision. Each frequency level captures progressively finer details, from diffuse colors and general structure to high-frequency variations, optimizing both the rendering process and frequency-aware modeling. 

To render a scene at a chosen frequency level, all Gaussians up to that frequency level are rendered in a single render pass. In particular, rendering the full-resolution image requires simply rendering all Gaussians in one pass, just as in standard 3D Gaussian splatting.

\subsection{Preliminary: 3D Gaussian Splatting}
3D Gaussian Splatting (3D-GS) \cite{kerbl20233d} represents a scene as a collection of anisotropic 3D Gaussians $\{G_i\}_{i=1}^N$, providing a structured, explicit alternative to neural-based representations.

The initialization of 3D-GS typically starts from a point cloud $\{x_i\}_{i=1}^N$ obtained via Structure-from-Motion (SfM). Each point $x_i$ in this cloud is assigned a 3D Gaussian representation $G_i$, which is defined by a mean position $\mu_i$ and a covariance matrix $\Sigma_i$. The covariance is decomposed as:

\begin{equation}
\Sigma_i = \mathcal{R S S^T R^T},
\label{eq:Covariance}
\end{equation}

where $\mathcal{S}$ represents a scaling matrix controlling the extent of the Gaussian, and $\mathcal{R}$ is a rotation matrix ensuring $\Sigma_i$ remains positive semi-definite. Additionally, each Gaussian stores appearance information through spherical harmonics and an opacity value $\alpha_i$, which contributes to the blending process.

Rendering in 3D-GS involves projecting the 3D Gaussians into 2D space, converting them into 2D Gaussians $G_i'$. A differentiable rasterization process sorts these 2D Gaussians by depth and blends them according to their opacity. The final color $C(x)$ at a pixel position $x$ is computed as:

\begin{equation}
C(x) = \sum_{j \in M} c_j \sigma_j \prod_{k=1}^{j-1} (1 - \sigma_k),
\label{eq:AlphaBlending}
\end{equation}

where $\sigma_j = \alpha_j G_j'(x)$, $M$ represents the ordered set of Gaussians contributing to $x$, and $c_j$ denotes the corresponding color. This differentiable rendering pipeline enables joint optimization of all Gaussian attributes during training, facilitating high-quality view synthesis.

\begin{figure}
    \centering
    \includegraphics[width=0.95\linewidth]{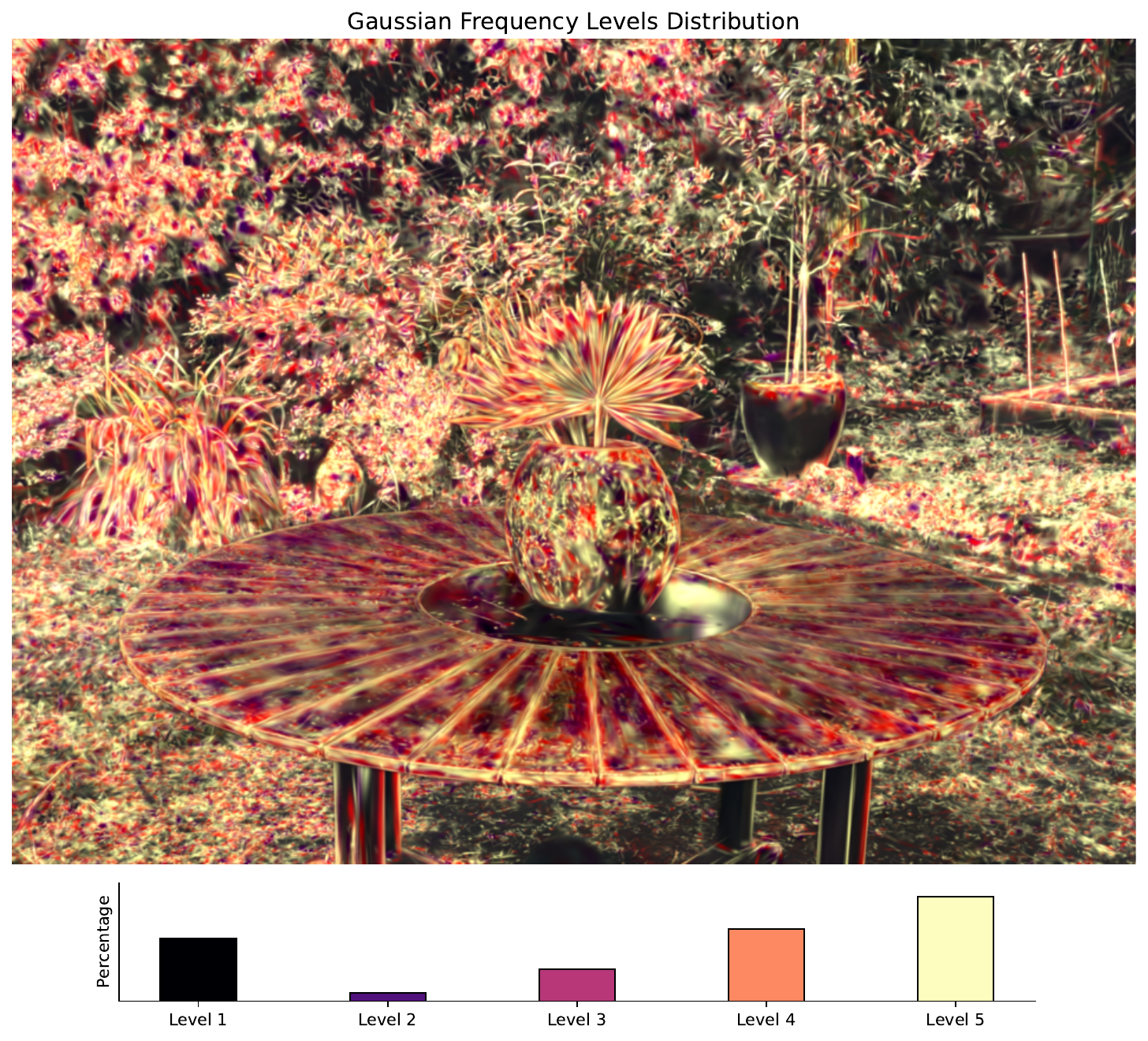}
    \caption{Visualization of the Gaussian frequency levels distribution in the scene. 
    The colormap represents different frequency levels, where lower levels correspond to coarse scene structures and higher levels capture finer details.}
    \vspace{-0.5em}
    \label{fig:gaussian_level_distribution}
\end{figure}

\subsection{Gaussian Level Grouping}

\cref{fig:short} gives an overview of our method. We progressively group 3D Gaussians based on their frequency contribution, ensuring that the accumulated rendering of the first $k$ groups matches a corresponding low-pass filtered version of the input. This guarantees that each individual level aligns with the respective frequency band in the Laplacian pyramid of the input images. To enforce frequency-aware consistency, we introduce a dedicated frequency regularization term in the loss function.

To hierarchically represent the scene by frequency levels, our method divides the Gaussians into groups, where each group corresponds to a specific band of frequencies. The scene is constructed progressively, starting from low frequencies and incorporating higher frequencies over time by adding additional groups. In practice, we modulate each Gaussian $G_i$ with an integer, non-learnable parameter $l_i$, termed "level", which denotes the frequency group to which the Gaussian belongs. We initialize and update this parameter as follows:

{\bf Initialization:} Gaussians of the first level are initialized using a Structure-from-Motion (SfM) point cloud, as in the original 3D Gaussian Splatting paper. All Gaussians are assigned an initial level of 1, representing the base frequency group.

{\bf Adaptive density control:} During training, Gaussians can be split or cloned using the adaptive density control technique from the original method. The newly created Gaussians inherit the same level as their parent.

{\bf Progressive Level Introduction:} Every $K$ training steps, we introduce a new level by duplicating all Gaussians and assigning them the next frequency level; almost all duplicates are removed during training (see Supplementary for Gaussian count dynamics).

\cref{fig:gaussian_level_distribution} shows the decomposition of 3D Gaussians to different groups that correspond to different levels of the Laplacian pyramid. As can be seen, the base level requires a considerable amount of Gaussians, while the intermediate low levels require far less, except for the last two levels, which correspond to high frequencies.

\subsection{Residual Color Gaussians}

Traditional Gaussian Splatting techniques rely on the accumulation of Gaussian colors to render scenes. However, in our method, higher-level Gaussians contribute details to lower-level Gaussians, such as textures and edges, which cannot be achieved efficiently by simply "adding color". To address this limitation, we introduce Residual Color Gaussians, allowing for both positive and negative color values. This approach compensates for lower-level contributions and facilitates the addition of high-frequency details, achieving a more accurate and detailed hierarchical representation.

Level 1 Gaussians retain colors in the range $[0,1]$, consistent with the original 3D-GS paper, by using view-dependent spherical harmonic color generation; Given a  Gaussian $G_i$ and viewing direction $d$, the rendered color of the Gaussian is computed as:
\begin{equation}
C_i=SH(G_i,d)
\label{eq:RegularSHToRGB}
\end{equation}
Inspired by the Laplacian pyramid \cite{burt1987laplacian} in image processing, where residual levels can take both positive and negative values to represent higher-frequency sub-bands, we multiply the color range of all Gaussians at higher levels by 2 and shift down by 1, resulting in a range of $[-1,1]$:
\begin{equation}
C_i=(SH(G_i, d) \cdot 2) - 1
\label{eq:NewSHToRGB}
\end{equation}
This simple adjustment is applied before alpha blending and enables subtraction of colors between positive and negative colored Gaussians. The final rendered color is clamped to the range $[0,1]$. This approach effectively captures high-frequency details, analogously to the functionality of residual levels in the Laplacian pyramid.

\subsection{Progressive Rendering of Frequency Bands}

As stated above, at inference time, one can render all Gaussians in the scene at once to get the full resolution rendered image.
Recall, in the image Laplacian Pyramid, one can reconstruct a Gaussian Pyramid by up-sampling and adding each Laplacian level with its successor, resulting in images with progressively finer details. Similarly, to regulate the different frequency levels in our method during the training stage, we render each level of the Gaussian Pyramid separately by combining all Gaussian groups up to the corresponding frequency level.

Importantly, all images are rendered at the same (full) resolution. This ensures that we can accurately regulate the frequencies of the Gaussians in each accumulated level without inadvertently removing valid high-frequency details due to subsampling. Subsampling would remove or alias frequency components above the Nyquist limit \cite{shannon1949communication}, making it impossible to determine whether Gaussians truly represent low-frequency information or if their details were simply lost during down-sampling. Rendering at full resolution allows us to preserve and regulate the true frequency content of each level accurately.

Specifically, for a pyramid with $L$ levels, rendering the $k$-th frequency level, where $1\leq k \leq L$, we render all Gaussians $G_i$ that satisfy $l_i \leq k$, resulting in the following sequence of rendered images:
\begin{equation}
\{\hat{I}_k\}_{k=1}^L, \quad \hat{I}_k \in \mathbb{R}^{H \times W \times 3}, \quad 1\leq k \leq L
\label{eq:LowPassFilterPyramidRendering}
\end{equation}

\subsection{Frequency Regulation and Losses Design}

As mentioned earlier, we apply frequency regulation at full resolution. To achieve this, we utilize bilinear interpolation with a scale factor of $0.5$ for downsampling, denoted as $\downarrow_2$, and bilinear interpolation with a scale factor of $2$ for upsampling, denoted as $\uparrow_2$.

Hence, given a GT image $I \in \mathbb{R}^{H \times W \times 3}$, in order to regulate the frequency band of the $k$-th level $\hat{I}_k$, we construct the low-pass filtered image with:
\begin{equation}
I_k=((I)\downarrow_2^{L-k})\uparrow_2^{L-k}
\label{eq:BlurGT}
\end{equation}

This way, we obtain the frequency spectrum of the GT image as it was subsampled $L-k$ times, but we keep it in its original resolution.

\begin{table*}[t]
\centering
\resizebox{\linewidth}{!}{%
\begin{tabular}{l|ccccc|ccccc|ccccc}
\toprule
\textbf{Method} &
\multicolumn{5}{c|}{\textbf{Mip-NeRF360}} &
\multicolumn{5}{c|}{\textbf{Tanks \& Temples}} &
\multicolumn{5}{c}{\textbf{Deep Blending}} \\
 & SSIM$\uparrow$ & PSNR$\uparrow$ & LPIPS$\downarrow$ & Gauss$\downarrow$ & FPS$\uparrow$
 & SSIM$\uparrow$ & PSNR$\uparrow$ & LPIPS$\downarrow$ & Gauss$\downarrow$ & FPS$\uparrow$
 & SSIM$\uparrow$ & PSNR$\uparrow$ & LPIPS$\downarrow$ & Gauss$\downarrow$ & FPS$\uparrow$ \\
\midrule
3DGS              & 0.822 & 27.76 & 0.177 & 2673263 & 124.4
                  & \cellcolor{yellow!25}0.849 & 23.73 & \cellcolor{yellow!25}0.171 & 1571073 & 120.6
                  & 0.903 & 29.75 & 0.241 & 2465386 & 89.8 \\

Lapis-GS**        & 0.753 & 26.32 & 0.248 & 6252724 & 62.7
                  & 0.811 & 23.27 & 0.215 & 2864263 & 63.70
                  & 0.892 & 28.94 & 0.222 & 3874530 & 48.1 \\

FreGS*            & 0.826 & 27.85 & 0.209 & -- & --
                  & \cellcolor{yellow!25}0.849 & 23.96 & 0.178 & -- & --
                  & \cellcolor{yellow!25}0.904 & 29.93 & 0.240 & -- & -- \\

FLOD              & \cellcolor{yellow!25}0.832 & \cellcolor{yellow!25}28.20 & 0.174 & 2320942 & 140.3
                  & 0.845 & \cellcolor{orange!25}24.27 & 0.188 & 1314046 & 142.2
                  & 0.898 & 29.24 & 0.253 & 1783253 & 110.2 \\


MSGS*             & --   & 27.39 & \cellcolor{red!25}0.155 & -- & 109.9
                  & --   & 23.46 & \cellcolor{red!25}0.111 & -- & 131.6
                  & --   & 29.70 & \cellcolor{red!25}0.096 & -- & 135.1 \\

OCTree-GS         & \cellcolor{yellow!25}0.832 & \cellcolor{yellow!25}28.20 & 0.168 & \cellcolor{red!25}363001 & \cellcolor{yellow!25}176.3
                  & 0.822 & 23.79 & 0.224 & \cellcolor{red!25}131974 & \cellcolor{orange!25}202.4
                  & \cellcolor{yellow!25}0.904 & \cellcolor{yellow!25}30.21 & 0.251 & \cellcolor{red!25}181506 & \cellcolor{yellow!25}174.4 \\

Ours 5LOD         & \cellcolor{orange!25}0.867 & \cellcolor{orange!25}29.29 & \cellcolor{yellow!25}0.166 & \cellcolor{yellow!25}940257 & \cellcolor{orange!25}190.6
                  & \cellcolor{orange!25}0.878 & \cellcolor{yellow!25}24.21 & \cellcolor{red!25}0.111 & \cellcolor{yellow!25}898358 & \cellcolor{yellow!25}177.1
                  & \cellcolor{orange!25}0.912 & \cellcolor{orange!25}30.29 & \cellcolor{orange!25}0.160 & \cellcolor{orange!25}704216 & \cellcolor{orange!25}181.0 \\

Ours 3LOD         & \cellcolor{red!25}0.874   & \cellcolor{red!25}29.65 & \cellcolor{orange!25}0.156 & 789182\cellcolor{orange!25} & \cellcolor{red!25}224.6
                  & \cellcolor{red!25}0.879   & \cellcolor{red!25}24.40 & \cellcolor{orange!25}0.120 & \cellcolor{orange!25}493681 & \cellcolor{red!25}241.3
                  & \cellcolor{red!25}0.914   & \cellcolor{red!25}30.30 & \cellcolor{yellow!25}0.173 & \cellcolor{yellow!25}707819 & \cellcolor{red!25}186.0 \\

\bottomrule
\end{tabular}}
\caption{Comparison between our method and competing methods across three datasets. Cell colors indicate first (red), second (orange), and third (yellow) place for each metric. Our method ranks in the top 3 across all metrics and datasets, achieving the best SSIM, PSNR, and FPS in every case.  
\textbf{*} Indicates results taken from the original paper, not re-evaluated locally.  
\textbf{**} Lapis-GS was trained with $\lambda_{\text{SSIM}} = 0.2$ following their suggestion to overcome GPU OOM (OOM on a 48GB Nvidia A6000).}
\label{tab:quantitative-results}
\end{table*}

To regulate the frequency spectrum, we introduce the frequency magnitude discrepancy term, which helps maintain the desired frequencies in $\hat{I}_k$. Given an image \( I_k \), let \( F_k(u, v) \) denote its Discrete Fourier Transform (DFT) at frequency coordinate \( (u, v) \). Using the frequency magnitudes \( |F_k(u, v)| \), we define the frequency magnitude discrepancy term between two images as:
\begin{equation}
d_{\text{DFT}}(I_k, \hat{I}_k) = \frac{1}{UV} \sum_{u=0}^{U-1} \sum_{v=0}^{V-1} \left\lVert |F_k(u, v)| - |\hat{F}_k(u, v)| \right\rVert
\label{eq:DFTDiscrepencyTerm}
\end{equation}
where \(UV\) represents the total number of frequency components and \( F_k(u,v) \) and \( \hat{F}_k(u, v) \) denote the frequency coefficients of images \( I_k \) and \( \hat{I}_k \), respectively.

Employing \cref{eq:DFTDiscrepencyTerm}, we will define our frequency magnitude loss by applying the frequency discrepancy term on all levels, except for the last one (as we do not desire to limit its frequency band):
\begin{equation}
\mathcal{L}_{\text{DFT}} = \sum_{k=1}^{L-1} d_{\text{DFT}}(I_k, \hat{I}_k)
\label{eq:DFTLoss}
\end{equation}

Furthermore, we adopt a spatial image discrepancy term similar to the one used in the original 3D-GS paper. Specifically, we regulate the \( k \)-th frequency level at its corresponding low resolution. To prevent aliasing, we employ the REDUCE operation, as originally introduced in the Laplacian Pyramid framework \cite{burt1987laplacian}.

Given an image \( I \) and a 2D Gaussian kernel \( \mathcal{G} \), the REDUCE operation is defined as a convolution with the kernel followed by subsampling every second row and column:

\begin{equation}
\text{REDUCE}(I) = (I * \mathcal{G})_{::2}
\label{eq:ReduceOp}
\end{equation}

To regulate the \( k \)-th frequency level, we apply \cref{eq:ReduceOp} \( L-k \) times to both \( I_k \) and \( \hat{I}_k \), yielding:

\begin{equation}
I'_k = \text{REDUCE}^{L-k}(I_k), \quad \hat{I}'_k = \text{REDUCE}^{L-k}(\hat{I}_k)
\label{eq:ApplyReduceImageLoss}
\end{equation}

Using the processed images from \cref{eq:ApplyReduceImageLoss}, we define our spatial image discrepancy term as:
\begin{equation}
d_{\text{IM}}(I_k, \hat{I}_k) = (1-\lambda_{\text{SSIM}}) |I'_k - \hat{I}'_k| + \lambda_{\text{SSIM}} \, d_{\text{D-SSIM}}(I'_k, \hat{I}'_k)
\label{eq:ImDiscrepency}
\end{equation}
where \( d_{\text{IM}} \) combines a pixel-wise difference term with a structural similarity (SSIM) discrepancy $d_{\text{D-SSIM}}$, weighted by \( \lambda_{\text{SSIM}} \).
Utilizing \cref{eq:ImDiscrepency}, the spatial image loss is computed by summing the spatial image discrepancy terms over the rendered and ground truth (GT) levels:
\begin{equation}
\mathcal{L}_{\text{IM}} = \sum_{k=1}^L \lambda_{d_k}d_{\text{IM}}(I_k, \hat{I}_k)
\label{eq:ImageLoss}
\end{equation}

Finally, taking \cref{eq:ImageLoss} and \cref{eq:DFTLoss}, our final loss function is defined as:
\begin{equation}
\mathcal{L} = \lambda_{\text{IM}} \mathcal{L}_{\text{IM}} + \lambda_{\text{DFT}} \mathcal{L}_{\text{DFT}}
\label{eq:FinalLoss}
\end{equation}
where \(\lambda_{\text{IM}}\) and \(\lambda_{\text{DFT}}\) are weighting factors balancing the contributions of the spatial and frequency terms, respectively.

\section{Results}

We evaluate our method across a comprehensive set of metrics and benchmarks, aiming to validate both reconstruction quality and real-time rendering efficiency. Specifically, we report SSIM, PSNR, LPIPS, Gaussian count, and FPS, and compare against prior 3D Gaussian Splatting methods that support level-of-detail (LOD) rendering. For fair comparison, we re-trained and evaluated baselines using the same evaluation protocol.
In addition, we provide per-level progressive rendering analysis and demonstrate several applications uniquely enabled by our frequency-aware representation, including streaming, foveated rendering, and promptable 3D focus.


\textbf{Data and Evaluations:} Following the 3D-GS dataset configuration, we evaluated our model across all nine scenes of the Mip-NeRF 360 dataset \cite{barron2022mip}, the Truck and Train scenes from the Tanks \& Temples dataset \cite{knapitsch2017tanks}, and the Playroom and Dr. Johnson scenes from the Deep Blending dataset \cite{hedman2018deep}. For evaluation, we adopt the 3D-GS protocol, selecting every eighth image as part of the test set.

\subsection{Implementation Details}

Our implementation is based on the NerfStudio SplatFacto model \cite{tancik2023nerfstudio}. To ensure stable optimization, we begin training with low-resolution images and progressively double the training resolution each time a new frequency level is introduced. A new frequency level is added every $K=2500$ steps. Upon adding a level, all optimizers and their schedulers are reinitialized, and Gaussian refinement is temporarily disabled for 300 steps. Throughout training, all levels are jointly optimized. 
Following 3D-GS, we set $\lambda_{\text{SSIM}}=0.2$. Additionally, we empirically set $\lambda_{d_k}=0.1$ for $k<L$, and $\lambda_{d_k}=1$ for $k=L$. The final loss function is weighted as $\lambda_{IM}=1$ and $\lambda_{DFT}=0.001$.

All scenes are trained for 30,000 steps, with Gaussian refinement enabled in our model for the entire training process. Optimization is performed using SplatFacto’s optimizers, and all experiments are conducted on a single NVIDIA RTX A5000 GPU.
For our model, we incorporate the view-dependent anti-aliased opacity term proposed in Mip-Splatting \cite{yu2024mip}.

\subsection{Comparisons with the State-of-the-Art}

We compare our method against recent 3D Gaussian Splatting techniques, including FLOD \cite{seo2025flodintegratingflexiblelevel}, OCTree-GS \cite{ren2024octree}, MSGS~\cite{yan2024multi}, FreGS~\cite{zhang2024fregs}, and Lapis-GS~\cite{shi2025lapisgslayeredprogressive3d}. While most of these methods support explicit level-of-detail (LOD) rendering, FreGS is included for its use of frequency-aware modeling, even though it does not offer progressive rendering. We also include 3DGS~\cite{kerbl20233d} as a baseline. Wherever possible, we re-trained or re-evaluated competing methods locally using the same evaluation protocol. 

We evaluated two variants of our method: \textbf{Ours 3LOD}, which uses three frequency levels, and \textbf{Ours 5LOD}, which uses five. Both are trained with the same budget of 30,000 steps. These configurations allow fine-grained trade-offs between quality and performance, and demonstrate the flexibility of our frequency decomposition.

As shown in Table~\ref{tab:quantitative-results}, both our 3LOD and 5LOD configurations consistently rank among the top three methods across all metrics and datasets. Notably, \textbf{Ours 3LOD achieves the highest SSIM, PSNR, and FPS in every dataset}, establishing it as the best performing method in terms of both reconstruction quality and rendering speed. Our models maintain highly competitive LPIPS scores despite using significantly fewer Gaussians than dense baselines like 3DGS and FLOD.





\subsection{Progressive Rendering}
\begin{figure}
    \centering
    \includegraphics[width=0.95\linewidth]{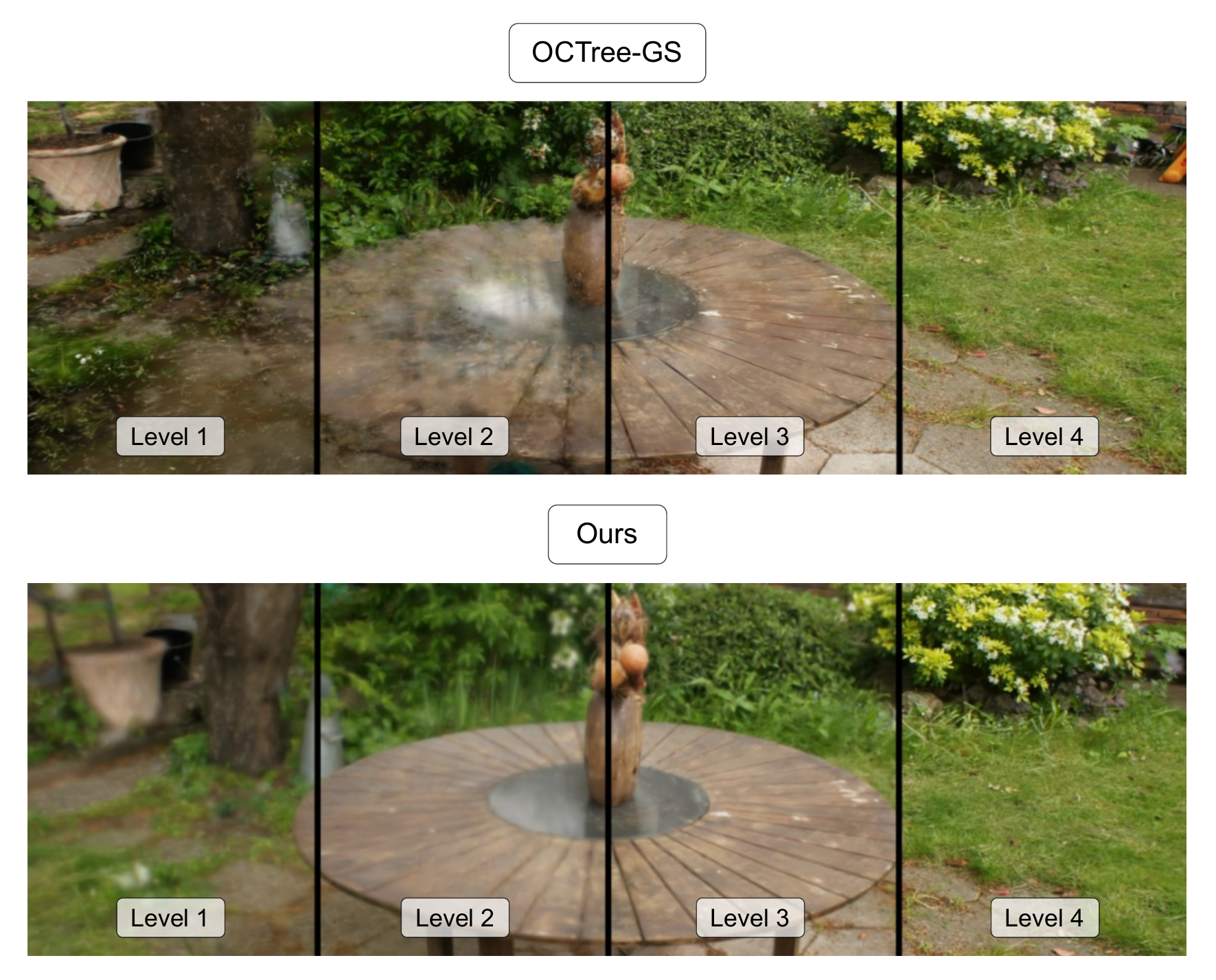}
    \caption{{\bf Progressive Rendering Consistency:} Comparing between OCTree-GS (top) and our method (bottom) across different frequency levels (left to right). In OCTree-GS, lower levels suffer from missing geometric details, which are recovered as more levels are added.
    In contrast, our representation preserves the scene’s geometry across all levels, with fine details progressively added, making it more suitable for frequency-aware applications.}
    \vspace{-0.5em}
    \label{fig:compare_octreegs}
\end{figure}

Progressive rendering on resource-constrained devices allows users to render only low-frequency levels for high frame rates, then progressively add higher-frequency details as computational budget allows. For streaming applications, low-frequency Gaussians can be transmitted first for immediate rendering, with higher-frequency content streamed progressively to refine quality over time.

Our frequency-based decomposition naturally enables progressive rendering where each level provides a meaningful, artifact-free approximation of the scene. Unlike spatial LOD methods that often introduce noticeable artifacts (Figure~\ref{fig:compare_octreegs}), our frequency hierarchy ensures smooth quality transitions and coherent approximations at all levels.

\begin{figure}
    \centering
    \includegraphics[width=0.95\linewidth]{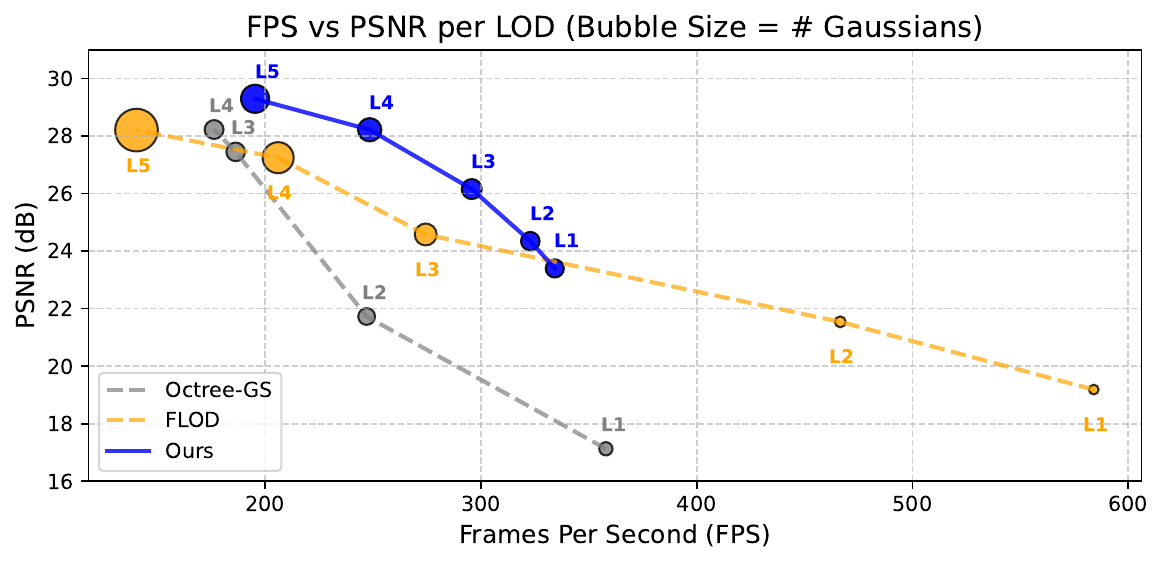}
    \caption{{\bf Progressive Rendering and Streaming:} Comparison of our method, FLOD, and OCTree-GS across different levels of detail. The plot shows the trade-off between rendering speed (FPS) and reconstruction quality (PSNR), where each point corresponds to a specific LOD level and bubble size reflects the number of Gaussians. Results are averaged over all Mip-NeRF 360 scenes. 
    Our method 
    demonstrating superior performance in both quality and speed in the highest three LODs. 
}
    \label{fig:psnr_vs_fps}
\end{figure}

As shown in Figure~\ref{fig:psnr_vs_fps}, our method consistently achieves the highest quality (PSNR) across all LODs while surpassing baselines in rendering speed from Level 3 onward.

\subsection{Faster Geometry Interaction}

\begin{figure*}[t]
    \centering
    %
    \begin{subfigure}[t]{0.355\linewidth}
        \centering
        \includegraphics[height=3.4cm,trim=12pt 12pt 12pt 12pt,clip]%
            {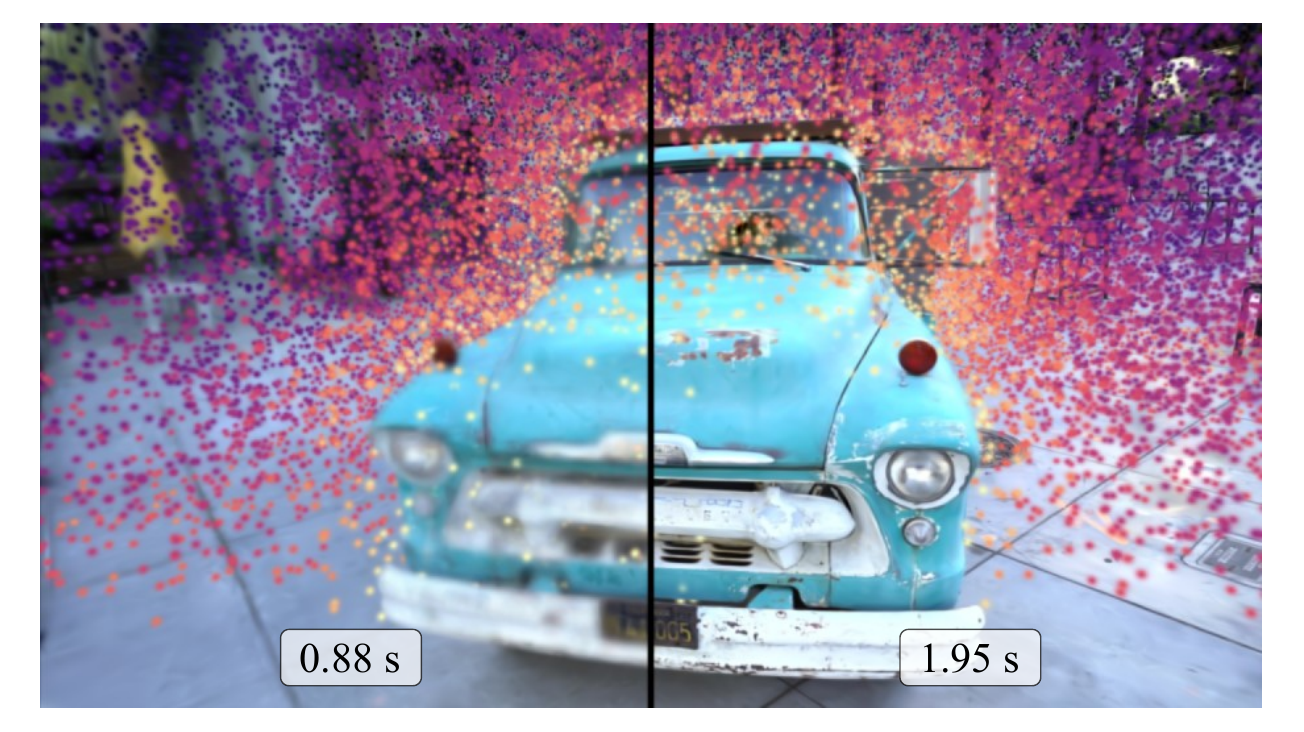}
        \caption{Efficient Geometry Interaction}
        \label{fig:collision_detection}
    \end{subfigure}
    \hspace{0.01\linewidth}
    %
    \begin{subfigure}[t]{0.3\linewidth}
        \centering
        \includegraphics[height=3.4cm,trim=0 0pt 0 0,clip]%
            {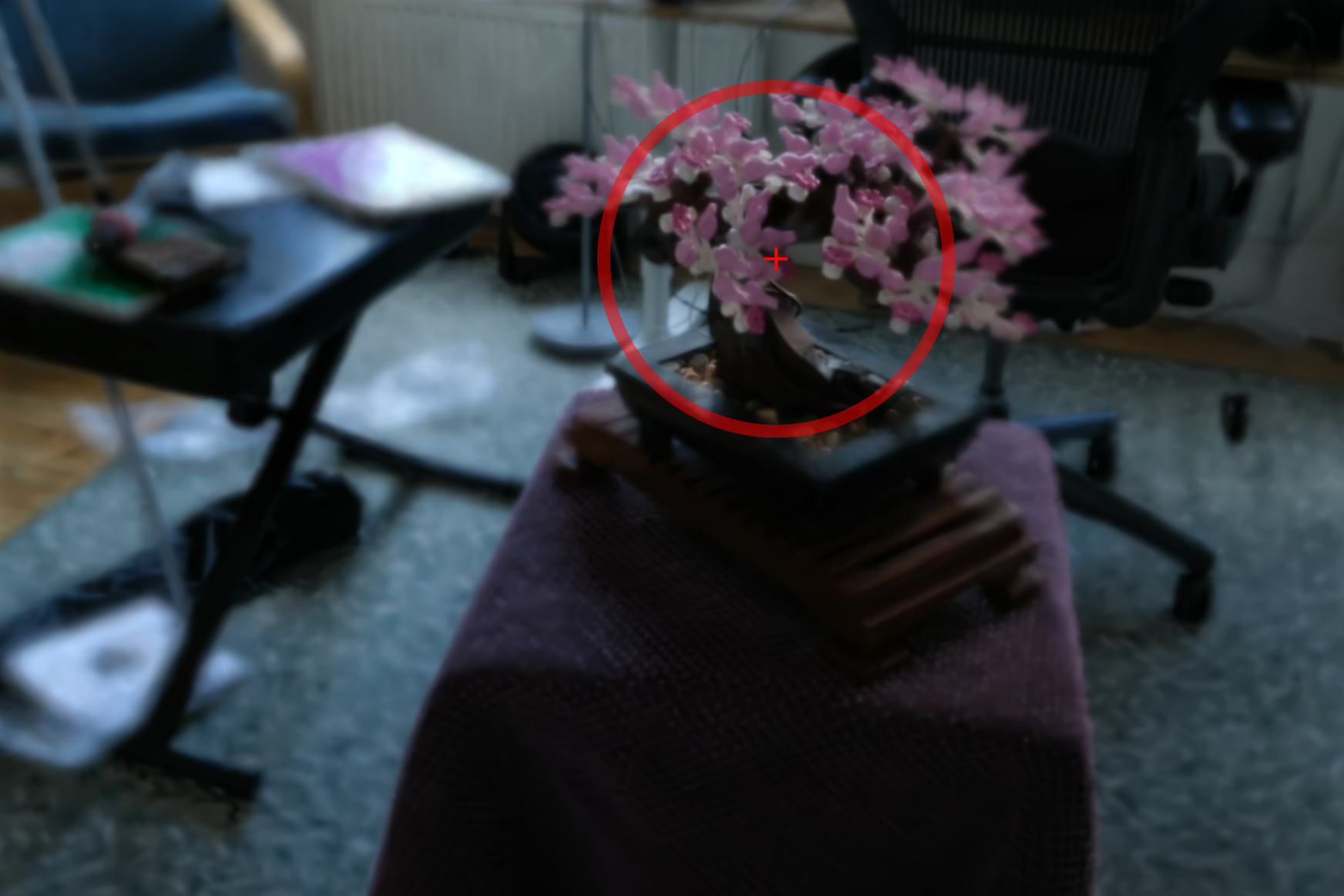}
        \caption{Foveated Rendering}
        \label{fig:foveated}
    \end{subfigure}
    \hspace{0.01\linewidth}
    %
    \begin{subfigure}[t]{0.3\linewidth}
        \centering
        \includegraphics[height=3.42cm,trim=0 6pt 0 10pt,clip]%
            {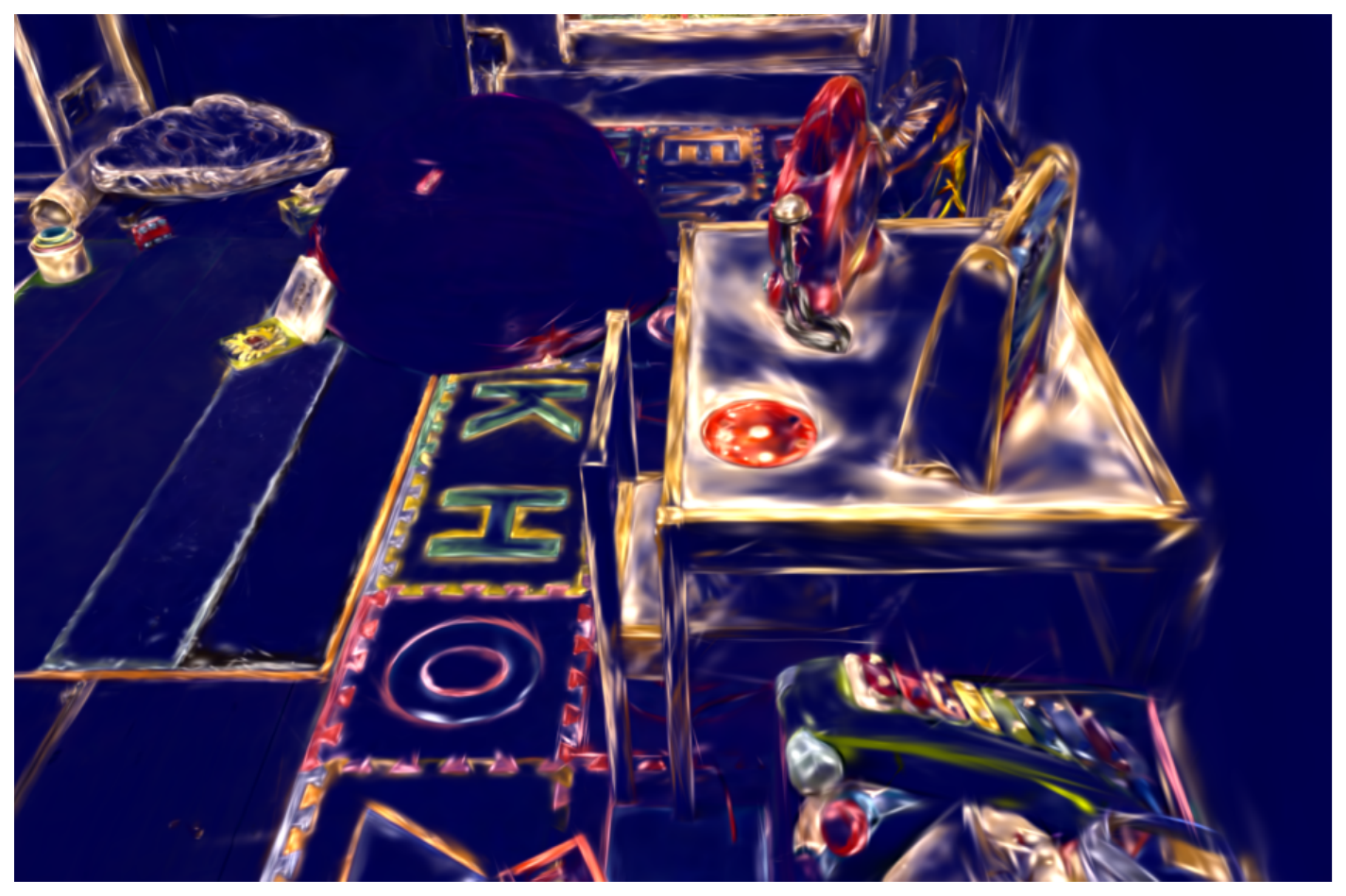}
        \caption{Artistic Visualization}
        \label{fig:xray_example}
    \end{subfigure}
    \caption{
      \textbf{Applications of Frequency-Based Gaussian Rendering:} (a) Efficient geometry interaction using low-frequency levels for faster processing. Closer points are brighter. Using only low frequency Gaussians (left) is more than twice as fast as using all Gaussians (right) with negligible error. (b) Foveated rendering, improving FPS performance using eye-tracking. (c) Frequency-based artistic effects, demonstrating an X-ray-like visualization. Please zoom in for details.
    }
    \label{fig:applications}
\end{figure*}

One important application of our frequency decomposition method is fast geometry interaction. Since low-frequency Gaussians define the general structure of the scene, many geometric operations can be efficiently approximated by operating only on this subset. Figure~\ref{fig:compare_octreegs} shows that we have structural integrity in lower levels.

We compare the performance of nearest-neighbor search on a randomly sampled 100K point cloud with a scale of 1, using either the full Gaussian set or only the low-frequency subset. Querying the KDTree \cite{bentley1975multidimensional} built from the full model required \textbf{1.95 seconds}, whereas querying the KDTree constructed from only the low-frequency Gaussians reduced this to \textbf{0.88 seconds}, achieving a \textbf{\(2.21\times\)} speedup. Despite this simplification, the median discrepancy with respect to the KDTree operating on the full model was \textbf{\(0.0006\)}, and the mean discrepancy was only \textbf{\(0.0018\)}, corresponding to a \textbf{\(0.18\%\)} error relative to the point cloud scale. This result suggests that many geometry-based interactions, such as collision detection and spatial querying, can be significantly accelerated using only the low-frequency representation, with minimal approximation error. The point cloud and the measured scene can be seen in \cref{fig:collision_detection}.

\subsection{Foveated Rendering}
Our representation enables efficient foveated rendering by dynamically adjusting the contribution of different frequency levels based on eye-tracking input, allowing faster rendering of the region of interest (ROI) relative to the entire image. This is particularly beneficial for AR/VR applications, where optimizing computational resources without sacrificing perceptual quality is critical. 

We modulate the contribution of 3D Gaussians at different frequency levels according to gaze-based weighting, discarding those whose effective opacity is below a threshold. As illustrated in Figure~\ref{fig:foveated}, this strategy preserves high-frequency detail in the area where the user is looking while smoothly reducing detail in peripheral regions. Leveraging our multi-resolution representation, we achieve a \textbf{40\% improvement in FPS} at 4K resolution, significantly reducing rendering costs without sacrificing perceptual fidelity. Further implementation details and additional examples are provided in the supplementary material.

\subsection{Promptable 3D Focus}
\begin{figure}
    \centering
    \includegraphics[width=0.95\linewidth]{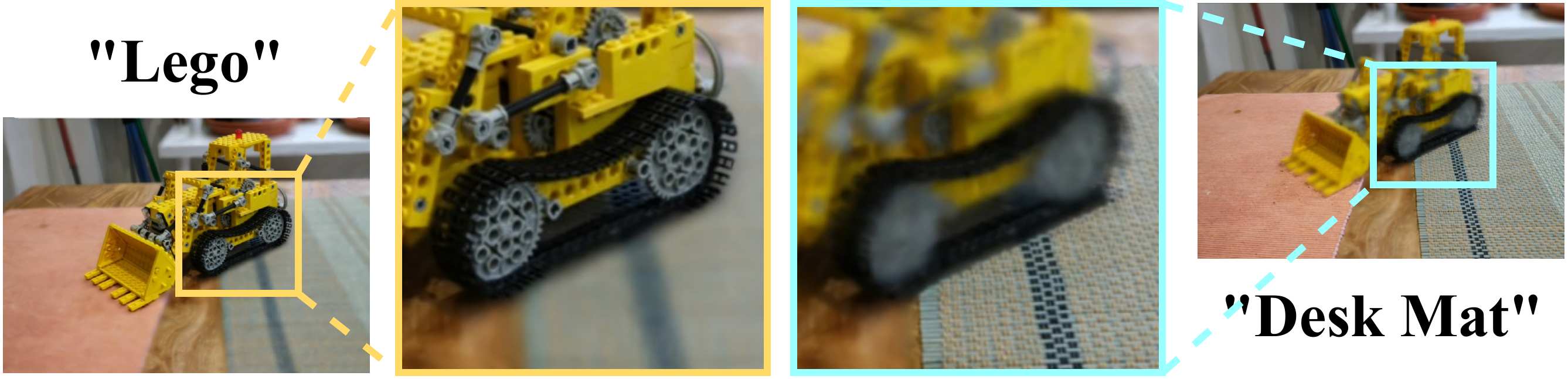}
    \caption{\textbf{Promptable 3D Focus:} Our method selectively renders only low-frequency Gaussians in the scene, except for Gaussians that belong to the prompted object. The left scene got the prompt "Lego", while the right scene got "Desk Mat".}
    \vspace{-0.5em}
    \label{fig:promptable_focus}
\end{figure}

Figure~\ref{fig:promptable_focus} demonstrates our text-prompted frequency-aware 3D focus, where users specify objects via natural language to achieve object-focused rendering that preserves high-frequency detail for targets while blurring the background. This overcomes limitations of 2D mask-based methods, that suffer from artifacts and poor multi-view consistency.

We leverage Lang-Segment-Anything \cite{medeiros2023langsam} for object identification. Given a text prompt, we generate masks across N training images, then identify corresponding Gaussians by projecting centers and computing mask overlap ratios. We preserve Gaussians that either (1) belong to the base frequency level or (2) exceed the mask threshold, producing 3D-consistent views with sharp objects and blurred backgrounds.

This approach offers both visual and efficiency gains: fewer Gaussians are rendered, enabling compact object-focused representations.


\subsection{Artistic 3D Filters}
Our representation also enables consistent 3D artistic filtering by selectively modifying the attributes of Gaussians at different frequency levels. By altering specific groups of Gaussians, we can achieve artistic effects that remain coherent across different viewpoints. To demonstrate that, one can use our representation to create an X-ray-style filter, where low-frequency Gaussians are assigned a constant color of dark blue to simulate an underlying structure, while high-frequency Gaussians have their RGB values increased by 0.2, enhancing the visibility of fine details. As demonstrated in Figure \ref{fig:xray_example}, this manipulation creates a 3D-consistent X-ray effect, preserving structural integrity across views.
Additional artistic effects on different scenes are showcased in the supplementary material.

\subsection{Ablation Studies}
To better understand the contributions of different components in our approach, we conducted a series of ablation studies. These experiments assess the impact of key design choices on model efficiency and rendering quality. All ablation studies were performed on the Kitchen scene from the Mip-NeRF 360 dataset, allowing for consistent evaluation across different configurations. Specifically, we examined (i) \textit{Residual Color Gaussians}, (ii) \textit{Image Loss on Lower Frequency Levels}, and (iii) \textit{Frequency Magnitude Loss}, revealing their respective influences on final rendering quality, memory requirements, and hierarchical decomposition. For a detailed discussion of each ablation, we refer the reader to the supplementary material.






\section{Conclusions}


We introduce frequency decomposition into 3D Gaussian Splatting, inspired by Laplacian pyramids in image processing, to provide structured control over frequency components in a scene. By organizing Gaussians into hierarchical subbands, incorporating Residual Color Gaussians, and leveraging progressive training, our method enhances interpretability and enables efficient level-of-detail management. 

Quantitatively, our method achieves state-of-the-art reconstruction quality and rendering speed among all level-of-detail extensions of 3D Gaussian Splatting, as demonstrated on standard benchmarks.
Additionally, this approach paves the way for applications such as promptable 3D focus and artistic filtering, allowing precise manipulation of frequency bands for creative effects and content-aware modifications. Furthermore, it supports practical scenarios like foveated rendering, progressive rendering, streaming, and fast geometry interaction, making real-time performance more adaptable across devices.  
These capabilities establish our method as both a versatile representation and a strong baseline for frequency-aware 3D scene modeling.


\begin{thebibliography}{39}
\providecommand{\natexlab}[1]{#1}
\providecommand{\url}[1]{\texttt{#1}}
\expandafter\ifx\csname urlstyle\endcsname\relax
  \providecommand{\doi}[1]{doi: #1}\else
  \providecommand{\doi}{doi: \begingroup \urlstyle{rm}\Url}\fi

\bibitem[Barron et~al.(2021)Barron, Mildenhall, Tancik, Hedman, Martin-Brualla, and Srinivasan]{barron2021mip}
Jonathan~T Barron, Ben Mildenhall, Matthew Tancik, Peter Hedman, Ricardo Martin-Brualla, and Pratul~P Srinivasan.
\newblock Mip-nerf: A multiscale representation for anti-aliasing neural radiance fields.
\newblock In \emph{Proceedings of the IEEE/CVF international conference on computer vision}, pages 5855--5864, 2021.

\bibitem[Barron et~al.(2022)Barron, Mildenhall, Verbin, Srinivasan, and Hedman]{barron2022mip}
Jonathan~T Barron, Ben Mildenhall, Dor Verbin, Pratul~P Srinivasan, and Peter Hedman.
\newblock Mip-nerf 360: Unbounded anti-aliased neural radiance fields.
\newblock In \emph{Proceedings of the IEEE/CVF conference on computer vision and pattern recognition}, pages 5470--5479, 2022.

\bibitem[Barron et~al.(2023)Barron, Mildenhall, Verbin, Srinivasan, and Hedman]{barron2023zip}
Jonathan~T Barron, Ben Mildenhall, Dor Verbin, Pratul~P Srinivasan, and Peter Hedman.
\newblock Zip-nerf: Anti-aliased grid-based neural radiance fields.
\newblock In \emph{Proceedings of the IEEE/CVF International Conference on Computer Vision}, pages 19697--19705, 2023.

\bibitem[Bentley(1975)]{bentley1975multidimensional}
Jon~Louis Bentley.
\newblock Multidimensional binary search trees used for associative searching.
\newblock \emph{Communications of the ACM}, 18\penalty0 (9):\penalty0 509--517, 1975.

\bibitem[Burt and Adelson(1987)]{burt1987laplacian}
Peter~J Burt and Edward~H Adelson.
\newblock The laplacian pyramid as a compact image code.
\newblock In \emph{Readings in computer vision}, pages 671--679. Elsevier, 1987.

\bibitem[Chen and Williams(2023)]{chen1993view}
Shenchang~Eric Chen and Lance Williams.
\newblock \emph{View Interpolation for Image Synthesis}.
\newblock Association for Computing Machinery, New York, NY, USA, 1 edition, 2023.

\bibitem[Hamdi et~al.(2024)Hamdi, Melas-Kyriazi, Mai, Qian, Liu, Vondrick, Ghanem, and Vedaldi]{hamdi2024ges}
Abdullah Hamdi, Luke Melas-Kyriazi, Jinjie Mai, Guocheng Qian, Ruoshi Liu, Carl Vondrick, Bernard Ghanem, and Andrea Vedaldi.
\newblock Ges: Generalized exponential splatting for efficient radiance field rendering.
\newblock In \emph{Proceedings of the IEEE/CVF Conference on Computer Vision and Pattern Recognition}, pages 19812--19822, 2024.

\bibitem[Hedman et~al.(2018)Hedman, Philip, Price, Frahm, Drettakis, and Brostow]{hedman2018deep}
Peter Hedman, Julien Philip, True Price, Jan-Michael Frahm, George Drettakis, and Gabriel Brostow.
\newblock Deep blending for free-viewpoint image-based rendering.
\newblock \emph{ACM Transactions on Graphics (ToG)}, 37\penalty0 (6):\penalty0 1--15, 2018.

\bibitem[Kerbl et~al.(2023)Kerbl, Kopanas, Leimk{\"u}hler, and Drettakis]{kerbl20233d}
Bernhard Kerbl, Georgios Kopanas, Thomas Leimk{\"u}hler, and George Drettakis.
\newblock 3d gaussian splatting for real-time radiance field rendering.
\newblock \emph{ACM Trans. Graph.}, 42\penalty0 (4):\penalty0 139--1, 2023.

\bibitem[Kerbl et~al.(2024)Kerbl, Meuleman, Kopanas, Wimmer, Lanvin, and Drettakis]{kerbl2024hierarchical}
Bernhard Kerbl, Andreas Meuleman, Georgios Kopanas, Michael Wimmer, Alexandre Lanvin, and George Drettakis.
\newblock A hierarchical 3d gaussian representation for real-time rendering of very large datasets.
\newblock \emph{ACM Transactions on Graphics (TOG)}, 43\penalty0 (4):\penalty0 1--15, 2024.

\bibitem[Khatib and Giryes(2024)]{khatib2024trinerflet}
Rajaei Khatib and Raja Giryes.
\newblock Trinerflet: A wavelet based triplane nerf representation.
\newblock In \emph{European Conference on Computer Vision}, pages 358--374. Springer, 2024.

\bibitem[Knapitsch et~al.(2017)Knapitsch, Park, Zhou, and Koltun]{knapitsch2017tanks}
Arno Knapitsch, Jaesik Park, Qian-Yi Zhou, and Vladlen Koltun.
\newblock Tanks and temples: Benchmarking large-scale scene reconstruction.
\newblock \emph{ACM Transactions on Graphics (ToG)}, 36\penalty0 (4):\penalty0 1--13, 2017.

\bibitem[Levoy and Hanrahan(1996)]{levoy1996light}
Marc Levoy and Pat Hanrahan.
\newblock Light field rendering.
\newblock page 31–42, New York, NY, USA, 1996. Association for Computing Machinery.

\bibitem[Liu et~al.(2024{\natexlab{a}})Liu, Luo, Fan, Wang, Peng, and Zhang]{liu2024citygaussian}
Yang Liu, Chuanchen Luo, Lue Fan, Naiyan Wang, Junran Peng, and Zhaoxiang Zhang.
\newblock Citygaussian: Real-time high-quality large-scale scene rendering with gaussians.
\newblock In \emph{European Conference on Computer Vision}, pages 265--282. Springer, 2024{\natexlab{a}}.

\bibitem[Liu et~al.(2024{\natexlab{b}})Liu, Luo, Mao, Peng, and Zhang]{liu2024citygaussianv2}
Yang Liu, Chuanchen Luo, Zhongkai Mao, Junran Peng, and Zhaoxiang Zhang.
\newblock Citygaussianv2: Efficient and geometrically accurate reconstruction for large-scale scenes.
\newblock \emph{arXiv preprint arXiv:2411.00771}, 2024{\natexlab{b}}.

\bibitem[Mallat(1989)]{mallat1989theory}
Stephane~G Mallat.
\newblock A theory for multiresolution signal decomposition: the wavelet representation.
\newblock \emph{IEEE transactions on pattern analysis and machine intelligence}, 11\penalty0 (7):\penalty0 674--693, 1989.

\bibitem[Meagher(1982)]{meagher1982geometric}
Donald Meagher.
\newblock Geometric modeling using octree encoding.
\newblock \emph{Computer graphics and image processing}, 19\penalty0 (2):\penalty0 129--147, 1982.

\bibitem[Medeiros et~al.(2023)]{medeiros2023langsam}
Luca Medeiros et~al.
\newblock Lang-segment-anything: Text-promptable segment anything model, 2023.
\newblock Accessed: 2025-02-01.

\bibitem[Mildenhall et~al.(2021)Mildenhall, Srinivasan, Tancik, Barron, Ramamoorthi, and Ng]{mildenhall2021nerf}
Ben Mildenhall, Pratul~P Srinivasan, Matthew Tancik, Jonathan~T Barron, Ravi Ramamoorthi, and Ren Ng.
\newblock Nerf: Representing scenes as neural radiance fields for view synthesis.
\newblock \emph{Communications of the ACM}, 65\penalty0 (1):\penalty0 99--106, 2021.

\bibitem[M{\"u}ller et~al.(2022)M{\"u}ller, Evans, Schied, and Keller]{muller2022instant}
Thomas M{\"u}ller, Alex Evans, Christoph Schied, and Alexander Keller.
\newblock Instant neural graphics primitives with a multiresolution hash encoding.
\newblock \emph{ACM transactions on graphics (TOG)}, 41\penalty0 (4):\penalty0 1--15, 2022.

\bibitem[Park et~al.(2017)Park, Yang, Yumer, Ceylan, and Berg]{park2017transformation}
Eunbyung Park, Jimei Yang, Ersin Yumer, Duygu Ceylan, and Alexander~C Berg.
\newblock Transformation-grounded image generation network for novel 3d view synthesis.
\newblock In \emph{Proceedings of the ieee conference on computer vision and pattern recognition}, pages 3500--3509, 2017.

\bibitem[Ren et~al.(2024)Ren, Jiang, Lu, Yu, Xu, Ni, and Dai]{ren2024octree}
Kerui Ren, Lihan Jiang, Tao Lu, Mulin Yu, Linning Xu, Zhangkai Ni, and Bo Dai.
\newblock Octree-gs: Towards consistent real-time rendering with lod-structured 3d gaussians.
\newblock \emph{arXiv preprint arXiv:2403.17898}, 2024.

\bibitem[Seo et~al.(2025)Seo, Choi, Son, and Uh]{seo2025flodintegratingflexiblelevel}
Yunji Seo, Young~Sun Choi, Hyun~Seung Son, and Youngjung Uh.
\newblock Flod: Integrating flexible level of detail into 3d gaussian splatting for customizable rendering, 2025.

\bibitem[Shade et~al.(1998)Shade, Gortler, He, and Szeliski]{shade1998layered}
Jonathan Shade, Steven Gortler, Li-wei He, and Richard Szeliski.
\newblock Layered depth images.
\newblock In \emph{Proceedings of the 25th annual conference on Computer graphics and interactive techniques}, pages 231--242, 1998.

\bibitem[Shannon(1949)]{shannon1949communication}
Claude~E Shannon.
\newblock Communication in the presence of noise.
\newblock \emph{Proceedings of the IRE}, 37\penalty0 (1):\penalty0 10--21, 1949.

\bibitem[Shao et~al.(2025)Shao, Qiao, Zhang, and Meng]{10916987}
Mingwen Shao, Yuanjian Qiao, Kai Zhang, and Lingzhuang Meng.
\newblock Frequency-aware uncertainty gaussian splatting for dynamic scene reconstruction.
\newblock \emph{IEEE Transactions on Visualization and Computer Graphics}, 31\penalty0 (5):\penalty0 3558--3568, 2025.

\bibitem[Shi et~al.(2025)Shi, Morin, Gasparini, and Ooi]{shi2025lapisgslayeredprogressive3d}
Yuang Shi, Géraldine Morin, Simone Gasparini, and Wei~Tsang Ooi.
\newblock Lapisgs: Layered progressive 3d gaussian splatting for adaptive streaming, 2025.

\bibitem[Tancik et~al.(2020)Tancik, Srinivasan, Mildenhall, Fridovich-Keil, Raghavan, Singhal, Ramamoorthi, Barron, and Ng]{tancik2020fourier}
Matthew Tancik, Pratul Srinivasan, Ben Mildenhall, Sara Fridovich-Keil, Nithin Raghavan, Utkarsh Singhal, Ravi Ramamoorthi, Jonathan Barron, and Ren Ng.
\newblock Fourier features let networks learn high frequency functions in low dimensional domains.
\newblock \emph{Advances in neural information processing systems}, 33:\penalty0 7537--7547, 2020.

\bibitem[Tancik et~al.(2023)Tancik, Weber, Ng, Li, Yi, Wang, Kristoffersen, Austin, Salahi, Ahuja, et~al.]{tancik2023nerfstudio}
Matthew Tancik, Ethan Weber, Evonne Ng, Ruilong Li, Brent Yi, Terrance Wang, Alexander Kristoffersen, Jake Austin, Kamyar Salahi, Abhik Ahuja, et~al.
\newblock Nerfstudio: A modular framework for neural radiance field development.
\newblock In \emph{ACM SIGGRAPH 2023 Conference Proceedings}, pages 1--12, 2023.

\bibitem[Wang and Xu(2024)]{wang2024pygs}
Zipeng Wang and Dan Xu.
\newblock Pygs: Large-scale scene representation with pyramidal 3d gaussian splatting.
\newblock \emph{arXiv preprint arXiv:2405.16829}, 2024.

\bibitem[Xu et~al.(2023)Xu, Zhan, Zhang, Yu, Zhang, Theobalt, Shao, and Lu]{xu2023wavenerf}
Muyu Xu, Fangneng Zhan, Jiahui Zhang, Yingchen Yu, Xiaoqin Zhang, Christian Theobalt, Ling Shao, and Shijian Lu.
\newblock Wavenerf: Wavelet-based generalizable neural radiance fields.
\newblock In \emph{Proceedings of the IEEE/CVF International Conference on Computer Vision}, pages 18195--18204, 2023.

\bibitem[Yan et~al.(2024)Yan, Low, Chen, and Lee]{yan2024multi}
Zhiwen Yan, Weng~Fei Low, Yu Chen, and Gim~Hee Lee.
\newblock Multi-scale 3d gaussian splatting for anti-aliased rendering.
\newblock In \emph{Proceedings of the IEEE/CVF Conference on Computer Vision and Pattern Recognition}, pages 20923--20931, 2024.

\bibitem[Yang et~al.(2022)Yang, Benaim, Jampani, Genova, Barron, Funkhouser, Hariharan, and Belongie]{yang2022polynomial}
Guandao Yang, Sagie Benaim, Varun Jampani, Kyle Genova, Jonathan Barron, Thomas Funkhouser, Bharath Hariharan, and Serge Belongie.
\newblock Polynomial neural fields for subband decomposition and manipulation.
\newblock \emph{Advances in Neural Information Processing Systems}, 35:\penalty0 4401--4415, 2022.

\bibitem[Yu et~al.(2024)Yu, Chen, Huang, Sattler, and Geiger]{yu2024mip}
Zehao Yu, Anpei Chen, Binbin Huang, Torsten Sattler, and Andreas Geiger.
\newblock Mip-splatting: Alias-free 3d gaussian splatting.
\newblock In \emph{Proceedings of the IEEE/CVF Conference on Computer Vision and Pattern Recognition}, pages 19447--19456, 2024.

\bibitem[Zeng et~al.(2025)Zeng, Wang, Ju, and Guan]{zeng2025frequency}
Zhaojie Zeng, Yuesong Wang, Lili Ju, and Tao Guan.
\newblock Frequency-aware density control via reparameterization for high-quality rendering of 3d gaussian splatting.
\newblock In \emph{Proceedings of the AAAI Conference on Artificial Intelligence}, pages 9833--9841, 2025.

\bibitem[Zhang et~al.(2024)Zhang, Zhan, Xu, Lu, and Xing]{zhang2024fregs}
Jiahui Zhang, Fangneng Zhan, Muyu Xu, Shijian Lu, and Eric Xing.
\newblock Fregs: 3d gaussian splatting with progressive frequency regularization.
\newblock In \emph{Proceedings of the IEEE/CVF Conference on Computer Vision and Pattern Recognition}, pages 21424--21433, 2024.

\bibitem[Zhang et~al.(2025)Zhang, Pan, Bao, Zhang, Xiang, Jiang, and Bao]{zhang2025lookcloser}
Xiaoyu Zhang, Weihong Pan, Chong Bao, Xiyu Zhang, Xiaojun Xiang, Hanqing Jiang, and Hujun Bao.
\newblock Lookcloser: Frequency-aware radiance field for tiny-detail scene.
\newblock In \emph{Proceedings of the Computer Vision and Pattern Recognition Conference}, pages 16122--16132, 2025.

\bibitem[Zhou et~al.(2016)Zhou, Tulsiani, Sun, Malik, and Efros]{zhou2016view}
Tinghui Zhou, Shubham Tulsiani, Weilun Sun, Jitendra Malik, and Alexei~A Efros.
\newblock View synthesis by appearance flow.
\newblock In \emph{European conference on computer vision}, pages 286--301. Springer, 2016.

\bibitem[Zhu et~al.(2023)Zhu, Zhu, Zhang, Zhu, Ma, and Cao]{zhu2023pyramid}
Junyu Zhu, Hao Zhu, Qi Zhang, Fang Zhu, Zhan Ma, and Xun Cao.
\newblock Pyramid nerf: Frequency guided fast radiance field optimization.
\newblock \emph{International Journal of Computer Vision}, 131\penalty0 (10):\penalty0 2649--2664, 2023.

\end{thebibliography}
{
    \small
    \bibliographystyle{ieeenat_fullname}

}
\clearpage
\setcounter{section}{0}
\setcounter{page}{1}
\maketitlesupplementary

\begin{figure}
    \centering
    \includegraphics[width=\linewidth]{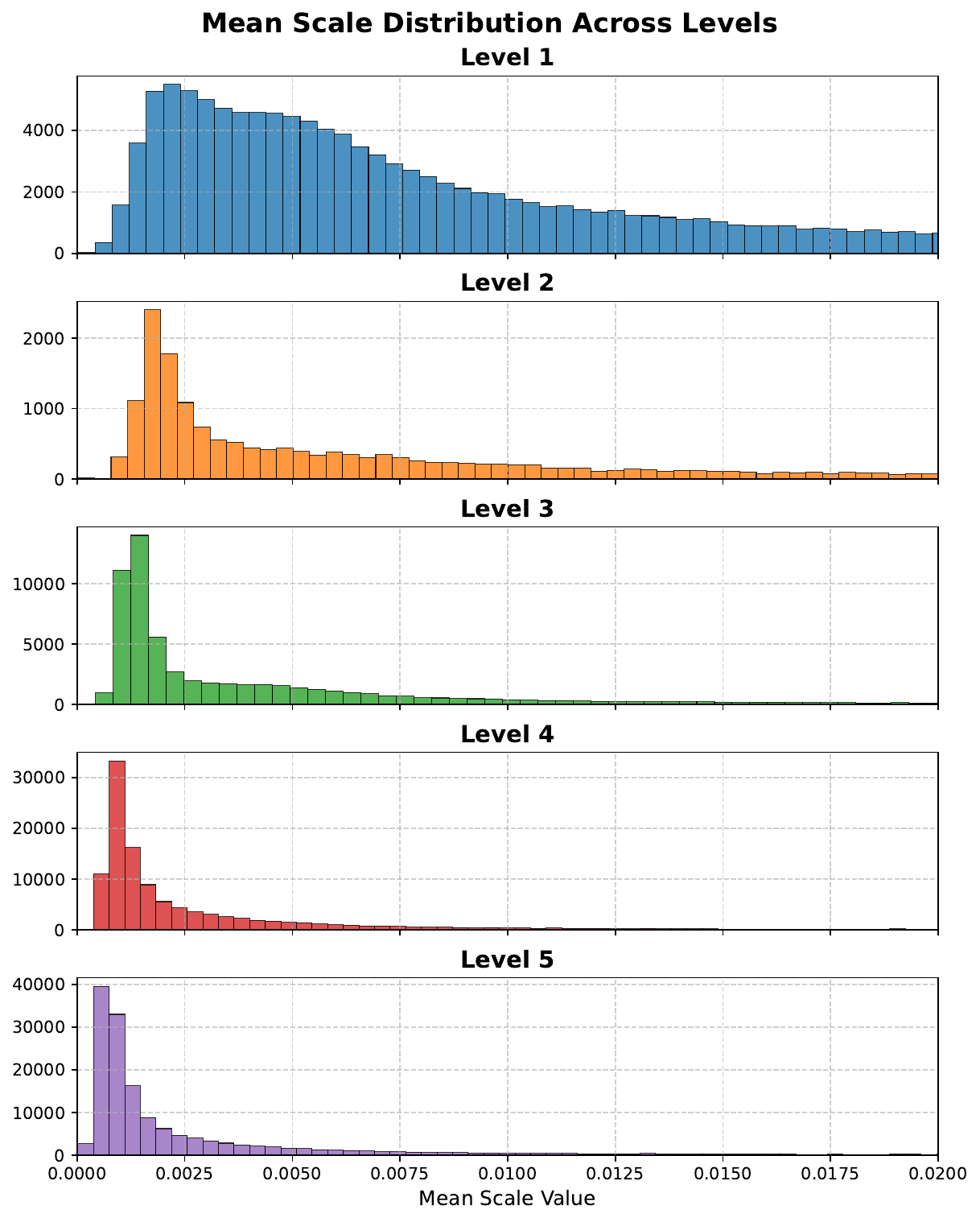}
    \caption{Mean scale distribution of Gaussians across different frequency levels. 
    As the frequency level increases, the distribution shifts toward smaller values, indicating a higher concentration of small Gaussians. This trend highlights the hierarchical nature of the representation, where lower levels capture coarse structures, and higher levels represent finer details.}
    \label{fig:mean_scale_distribution}
\end{figure}

\begin{figure}
    \centering
    \includegraphics[width=\linewidth]{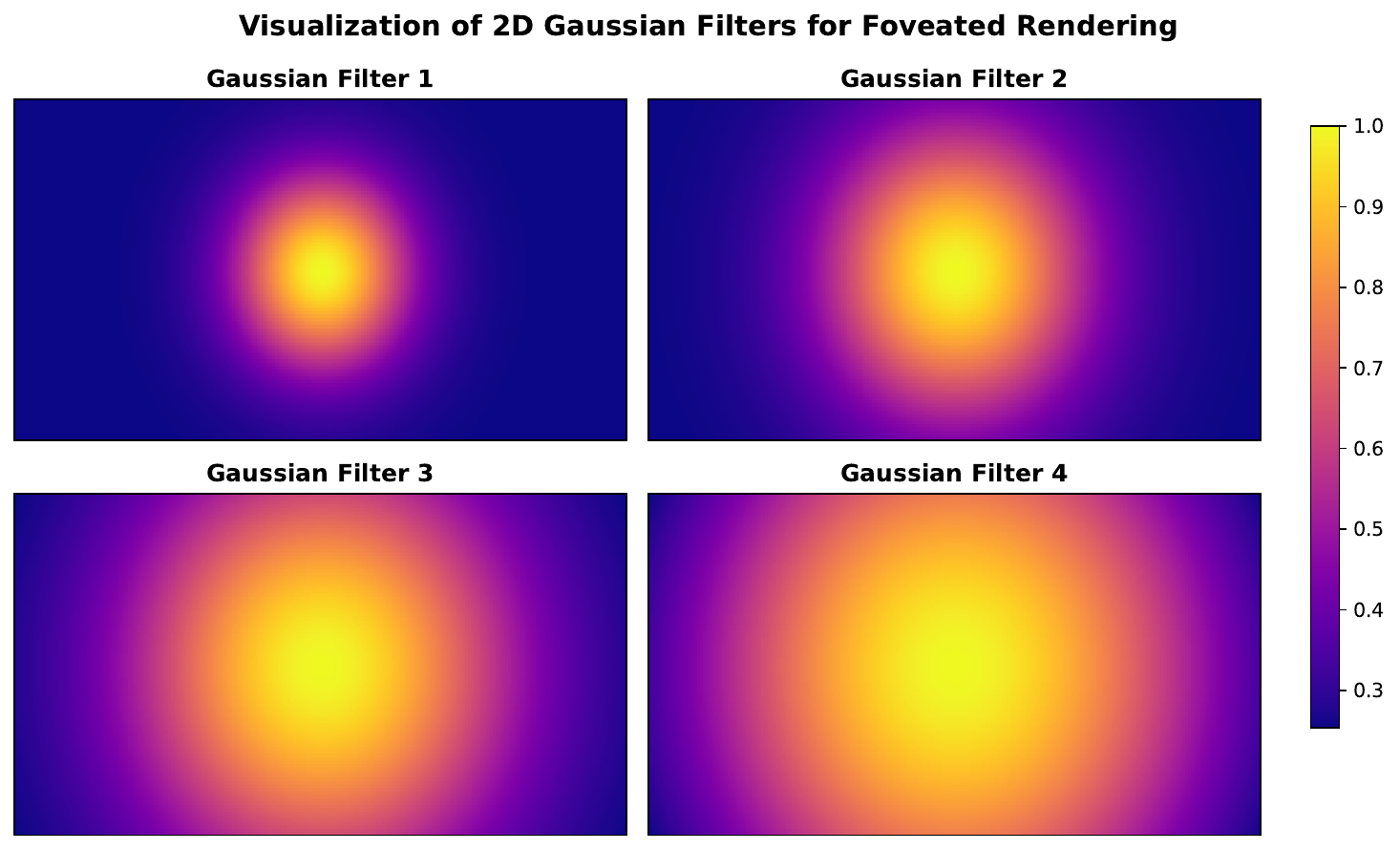}
    \caption{\textbf{Gaussian Filtering for Foveated Rendering:} Visualization of the hierarchical 2D Gaussian filters applied to projected 3D Gaussians at different frequency levels. Each filter progressively increases in standard deviation, ensuring that peripheral regions are rendered at lower fidelity while preserving details in the ROI.}
    \label{fig:gaussian_filters}
\end{figure}

\section{Supplementary Material}

This supplementary document provides additional details, experiments, and qualitative results to complement our main paper. Below is an overview of the contents:

\begin{itemize}
    \item \textbf{1.1 Analysis of Gaussian Scale Distribution Across Levels:} Distribution analysis of Gaussian scales across frequency levels.
    \item \textbf{1.2 Foveated Rendering:} Implementation details of our foveated rendering approach.
    \item \textbf{1.3 Ablation Studies:} Detailed discussion on key design choices, including residual color Gaussians, frequency magnitude loss, and image loss at lower levels.
    \item \textbf{1.4 Gaussian Count Dynamics:} Analysis of how the number of Gaussians evolves throughout training and across frequency levels.
    \item \textbf{1.5 Additional Examples of Level of Detail:} Visualization of level-of-detail decomposition in various scenes.
    \item \textbf{1.6 Additional Examples of Foveated Rendering:} Demonstrations of foveated rendering applied to different scenes.
    \item \textbf{1.7 Additional Examples of Promptable 3D Focus:} Illustration of selective object emphasis using our promptable 3D focus application in different scenes. 
    \item \textbf{1.8 Additional 3D Artistic Filters:} Examples of artistic filters applied to different scenes using our method.
\end{itemize}

\noindent The following sections provide in-depth discussions and results for each of these topics.

\subsection{Analysis of Gaussian Scale Distribution Across Levels}
To gain deeper insights into the characteristics of each frequency level in our representation, we analyze the distribution of the mean scale of the Gaussians, computed as the average of the scale values along the x, y, and z axes for each Gaussian. \Cref{fig:mean_scale_distribution} presents this distribution across different levels.

From the figure, we observe a clear trend: as the frequency level increases, the distribution of mean scales shifts toward smaller values, indicating a higher concentration of small Gaussians. This suggests that higher frequency levels predominantly capture fine details in the scene, represented by numerous small Gaussians, whereas lower frequency levels retain larger-scale structures. The progressive shift in the distribution reinforces the hierarchical nature of our representation, where coarse details are preserved in the lower levels, and finer structures emerge at higher levels.

\subsection{Foveated Rendering}
To implement foveated rendering, we apply a set of spatially varying 2D Gaussian filters (centered at the fixation point \((x, y)\)) to modulate Gaussians across our frequency levels. Each filter has an appropriately scaled standard deviation to retain peripheral lower-frequency components while prioritizing higher-frequency details in the ROI.

As depicted in Figure~\ref{fig:gaussian_filters}, these filters serve as soft importance weights for Gaussian opacities. Gaussians with contributions below a predefined threshold are discarded, effectively reducing computational load while maintaining visually coherent, high-detail renderings at fixation. This multi-scale filtering yields a notable 40\% FPS improvement at 4K resolution, underscoring its effectiveness for real-time AR/VR applications requiring both efficiency and high visual quality.

\subsection{Ablation Studies}
\textbf{Effect of Residual Color Gaussians.} We evaluated the effectiveness of Residual Color Gaussians by training a model where all Gaussians retained colors in the standard $[0,1]$ range, rather than using the proposed $[-1,1]$ residual representation for higher-level Gaussians. This modification led to a significant increase in the total number of Gaussians, from 409K to 515K, representing a 20\% growth in scene size. While this increase contributed to a slight improvement in PSNR (+0.2 dB), it also resulted in higher memory consumption and slower rendering times. These findings suggest that the residual color representation enables more compact and efficient scene encoding by allowing individual Gaussians to contribute more effectively to high-frequency details. The ability to add or remove color at different hierarchical levels grants each Gaussian greater expressive power, reducing the overall number of Gaussians required for accurate reconstruction.

\textbf{Effect of Image Loss on Lower Frequency Levels.} To assess the importance of applying the image loss at lower frequency levels, we trained a three-level model while omitting the image loss at lower levels. The resulting model achieved a comparable PSNR to our full method when evaluating the final rendered image; however, its intermediate frequency levels exhibited significant degradation. Specifically, when rendering downscaled resolutions using only the first level (corresponding to a $\times4$ downscaled output) or the first two levels ($\times2$ downscaled output), we observed a substantial loss of detail, leading to smeared and blurry reconstructions. Quantitatively, our full model achieved a PSNR of 29.65 dB for $\times4$ downscaled rendering and 32.27 dB for $\times2$ downscaled rendering. In contrast, the ablated model, trained without image loss at lower frequency levels, suffered a notable performance drop, with PSNR values of 25.97 dB and 29.36 dB for $\times4$ and $\times2$ downscaled renderings, respectively. These results highlight the crucial role of enforcing spatial consistency across frequency levels, ensuring that intermediate levels remain faithful to their expected resolution.

\textbf{Effect of Frequency Magnitude Loss.} We analyzed the impact of the frequency magnitude loss by training the same model without this term. Without frequency regulation, almost all scene details were modeled by the lowest-level Gaussians, while the higher levels contributed almost nothing. This disrupted the intended hierarchical decomposition, as the frequencies were not gradually distributed across levels but instead collapsed into the first level. Consequently, the model failed to leverage the benefits of multi-scale representation, relying predominantly on a single frequency band.

\subsection{Gaussian Count Dynamics}
\begin{figure}
    \centering
    \includegraphics[width=\linewidth]{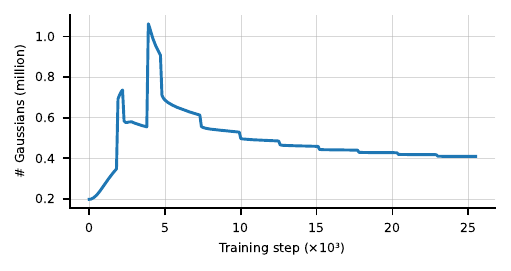}
    \caption{Evolution of the number of Gaussians during training. Each new frequency level increases the number of Gaussians, but adaptive density control quickly prunes low-utility ones, so the model converges to a compact, lightweight representation.}
    \label{fig:num_of_gaussians_vs_training_step}
\end{figure}

Figure~\ref{fig:num_of_gaussians_vs_training_step} illustrates the evolution of the number of Gaussians during training. Although each new frequency level initially doubles the number of Gaussians, the adaptive density control remains active throughout the entire training process, continuously removing low-utility components. As a result, the total number of Gaussians quickly stabilizes after each addition, resulting in a compact and efficient representation. This dynamic ensures that the final model remains lightweight despite the hierarchical structure.

\subsection{Additional Examples of Level of Detail}

\begin{figure*}[h]
\begin{center}
    \includegraphics[width=\textwidth]{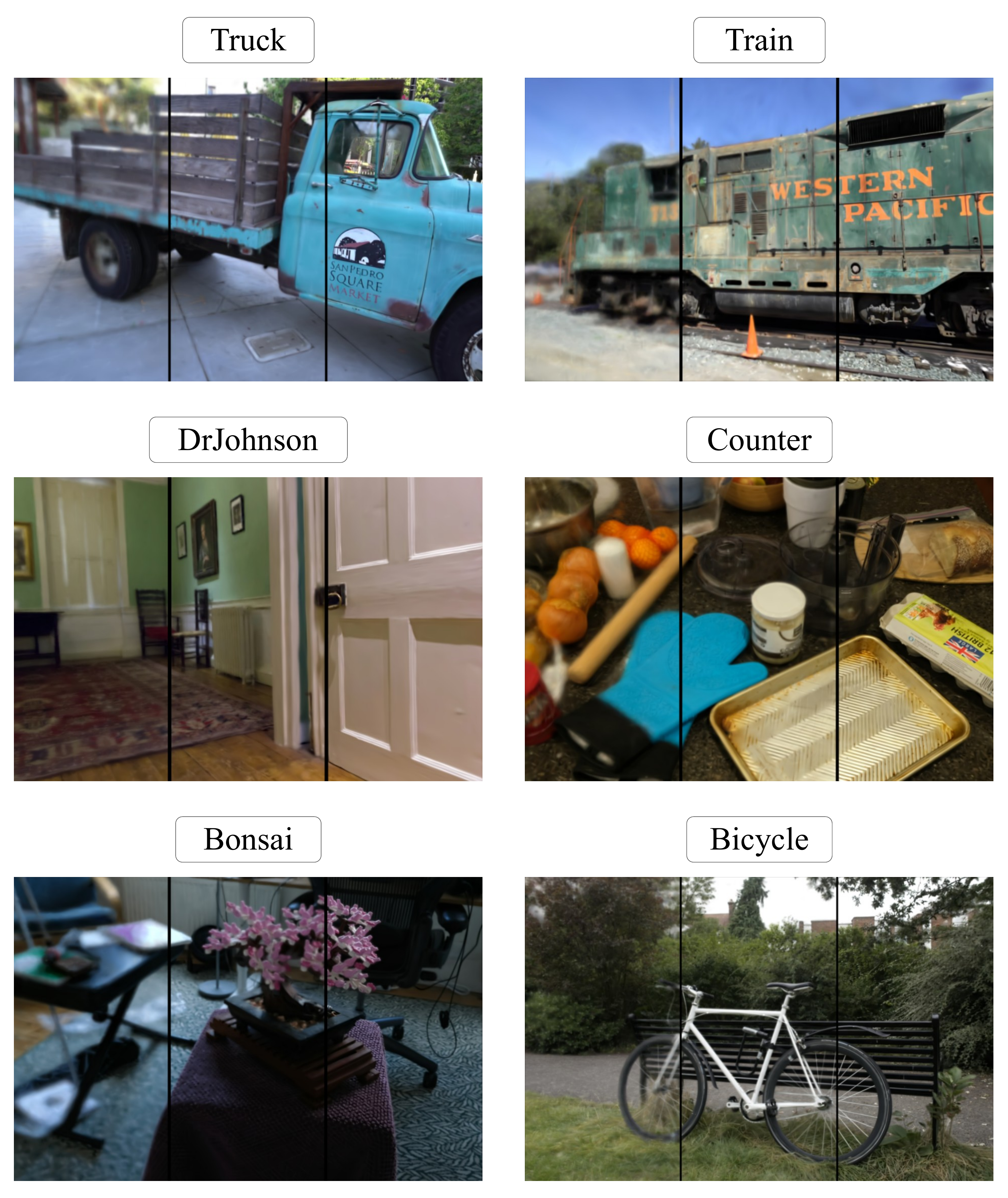}
\end{center}
\caption{Visualization of the hierarchical detail representation across various scenes using our method. Each image is segmented to illustrate different levels of structural and textural detail captured by our approach. The selected examples—\textit{Truck}, \textit{Train}, \textit{Dr. Johnson}, \textit{Counter}, \textit{Bonsai}, and \textit{Bicycle}—demonstrate its effectiveness across diverse real-world scenarios. Our model is trained with 5 levels, and we render levels 1, 3, and 5 for comparison.}
\label{fig:lod_examples}
\end{figure*}

In this section, we present qualitative results demonstrating the level of detail captured by our method across various scenes. Figure~\ref{fig:lod_examples} showcases six different scenarios, each divided into multiple segments, representing different levels of detail in our method’s representation. The images illustrate how our approach captures both coarse structural information and fine details, effectively adapting to the complexity of each scene. The selected examples include objects with varying textures, colors, and geometries, highlighting the versatility of our approach.

\subsection{Additional Examples of Foveated Rendering}

\begin{figure*}[t]
    \centering
    \includegraphics[width=0.9\linewidth]{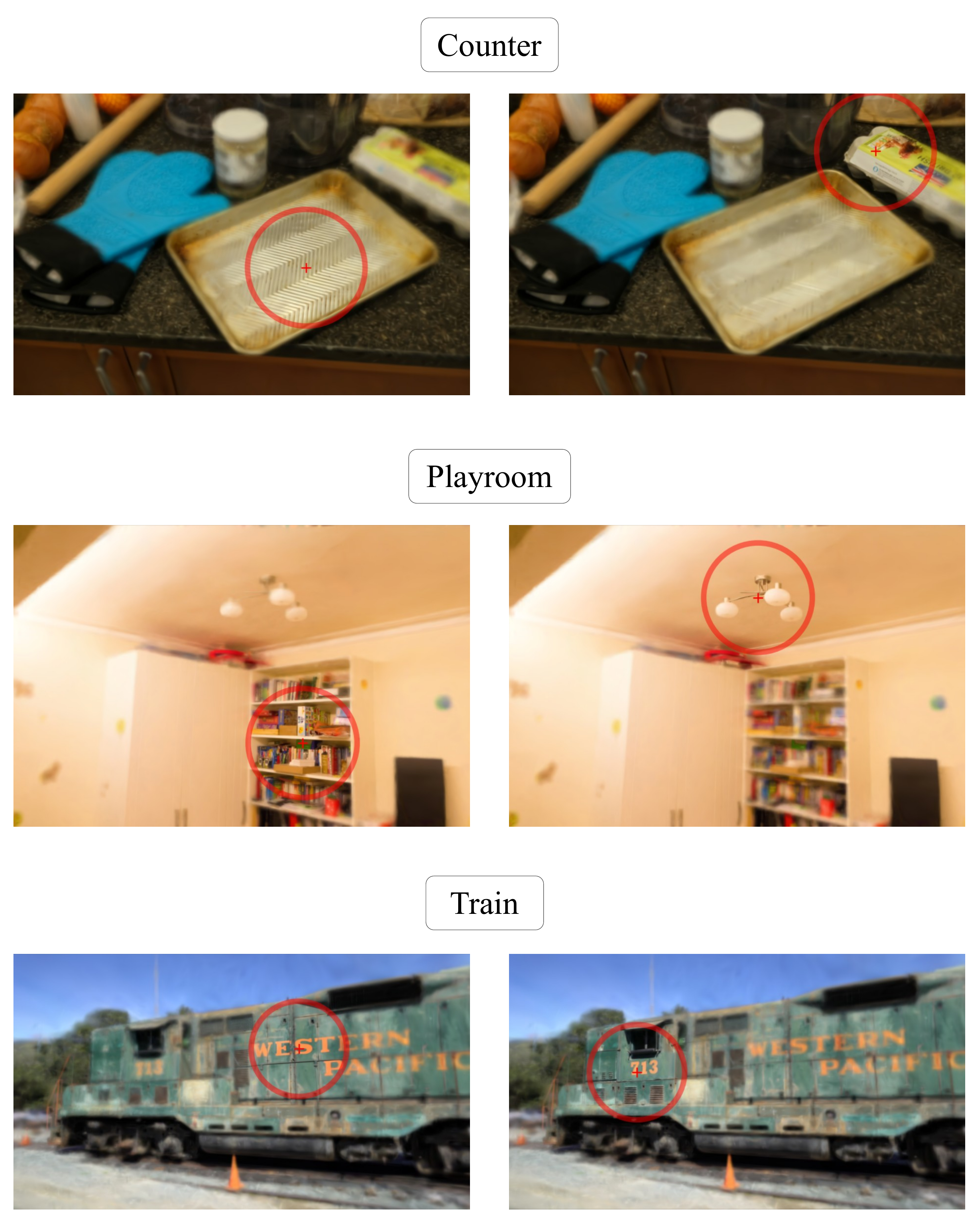}
    \caption{Foveated rendering applied to three different scenes: Counter, Playroom, and Train. Each scene contains two renderings with foveation applied to different areas (highlighted in red). The method prioritizes high-detail rendering in regions of interest while reducing detail elsewhere, optimizing rendering efficiency.}
\label{fig:foveated_rendering_more_examples}
\end{figure*}

In this section, we showcase the application of our foveated rendering method across three different scenes: Counter, Playroom, and Train. Each scene contains two renderings for comparison, where the foveation is applied at different regions. The method selectively enhances details in areas of interest while reducing computational load in less critical regions, effectively balancing visual quality and performance. \cref{fig:foveated_rendering_more_examples} demonstrates how our approach preserves important details in high-attention areas while applying aggressive downsampling elsewhere.

\subsection{Additional Examples of Promptable 3D Focus}

\begin{figure*}[t]
    \centering
    \includegraphics[width=0.9\linewidth]{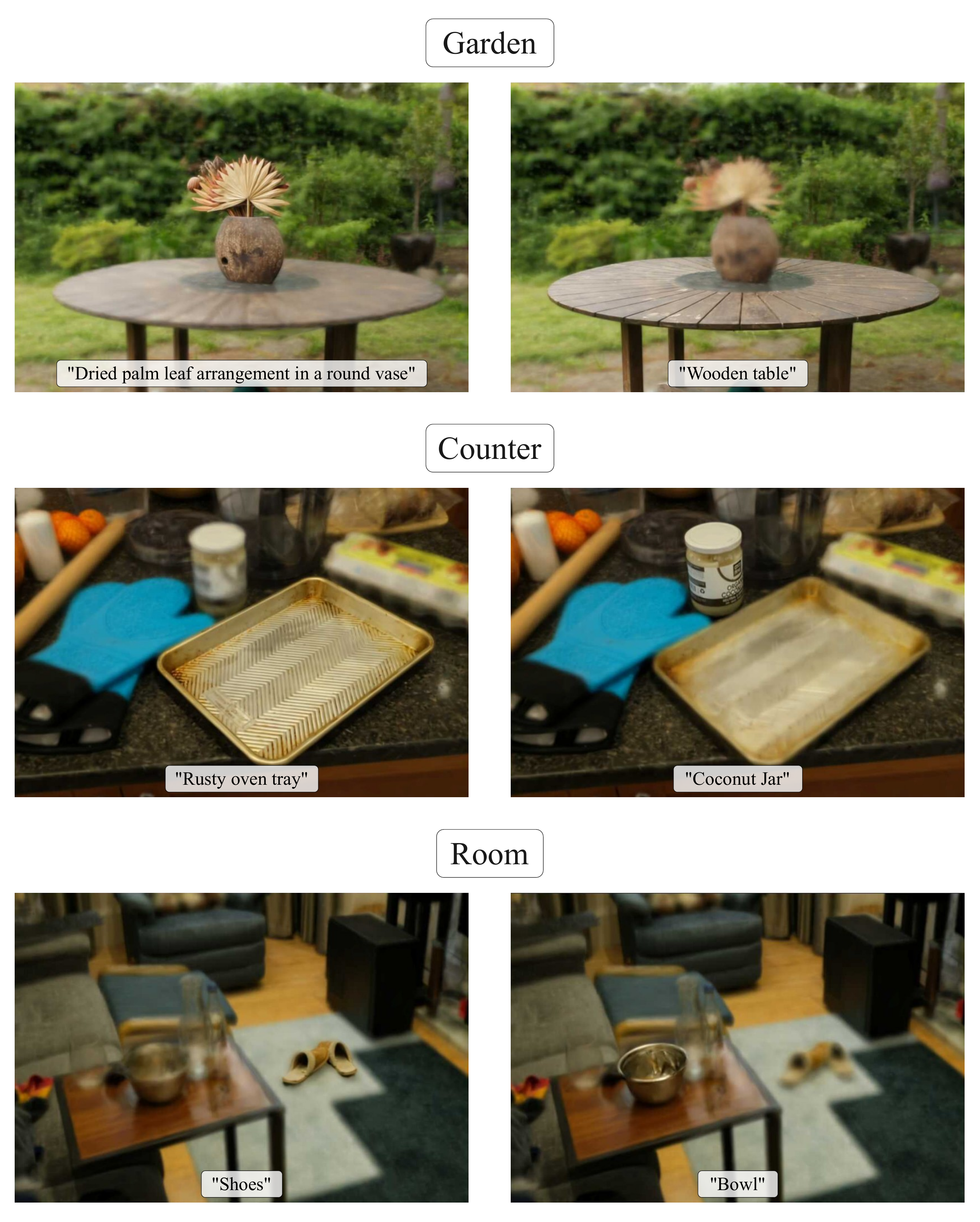}
    \caption{Additional results of our promptable 3D focus application. Each row shows a different object selected using a text prompt, demonstrating our method's ability to produce consistent 3D object emphasis.}
    \label{fig:editing_more_examples}
\end{figure*}

To further demonstrate the effectiveness of our promptable 3D focus application, we provide additional examples in Figure~\ref{fig:editing_more_examples}. These results highlight how our method enables selective object emphasis while maintaining 3D consistency. Each row in the figure presents different objects highlighted from a scene using text-based prompts, showcasing the robustness of our 3D voting mechanism. The approach effectively suppresses background details while preserving high-frequency structures for the selected objects, enabling seamless editing without introducing segmentation artifacts.

\subsection{Additional 3D Artistic Filters}
\begin{figure*}[t]
    \centering
    \includegraphics[width=\linewidth]{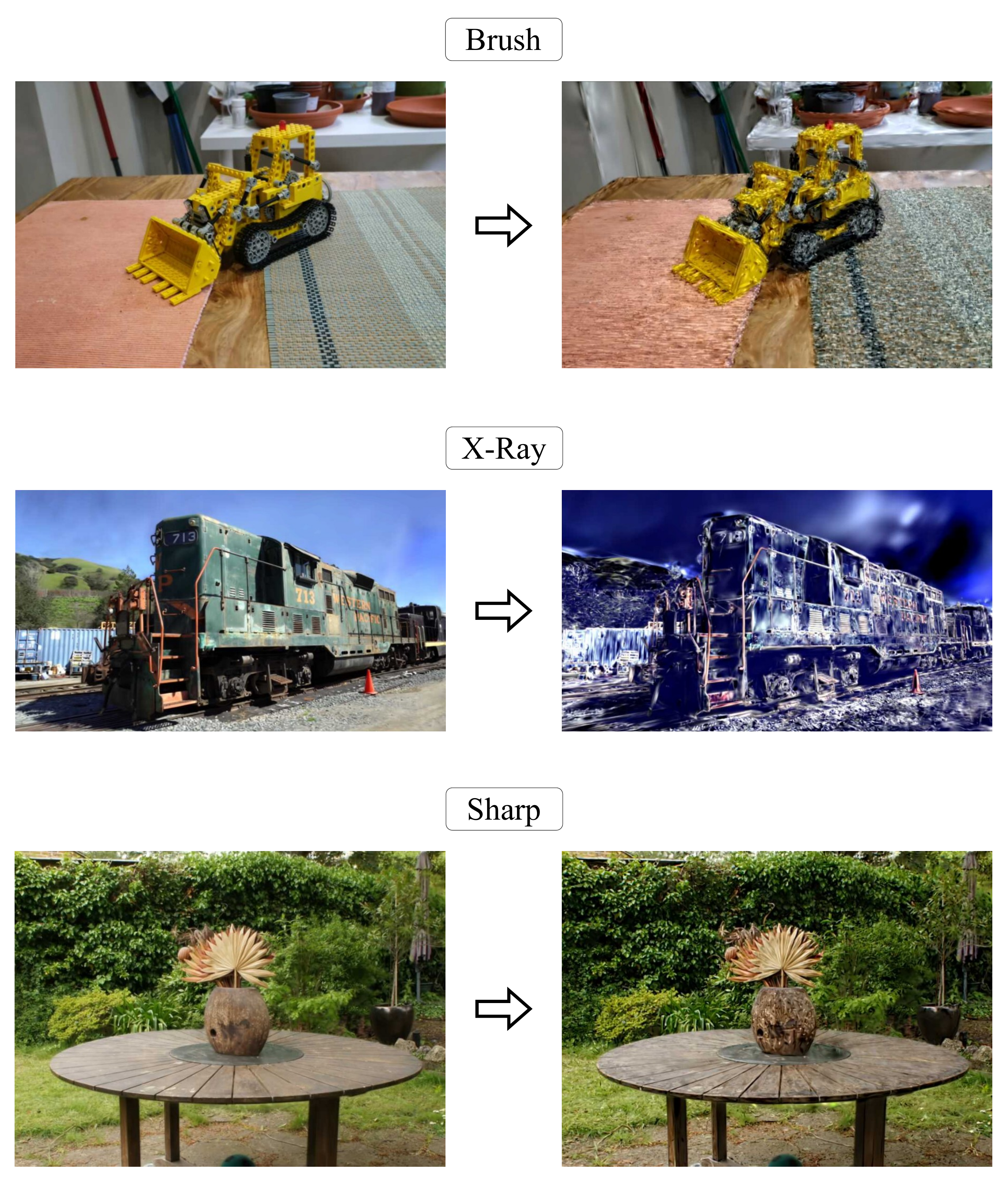}
    \caption{Three artistic effects applied to different scenes using our method. \textbf{Top:} The ``Brush'' effect introduces texture and painterly distortions by adjusting frequency bands and adding positional noise. \textbf{Middle:} The ``X-ray'' effect highlights high-frequency details while suppressing lower levels to create a glowing structural representation. \textbf{Bottom:} The ``Sharp'' effect enhances fine details by removing mid-frequency Gaussians and fully opacifying higher levels.}
    \label{fig:artistic_filters_more_examples}
\end{figure*}

We present more examples of artistic filters applied to different scenes using our method, as shown in Figure~\ref{fig:artistic_filters_more_examples}. These effects are achieved by selectively modifying the attributes of Gaussians at different frequency levels.

For the \textbf{Brush} effect, we modify the 5-level trained model by keeping level 1 unchanged, while brightening levels 2 and 4 and darkening levels 3 and 5. Additionally, we introduce Gaussian noise to the center positions of Gaussians at levels 2 through 5, resulting in a painterly, textured appearance.

For the \textbf{X-ray} effect, we set the colors of Gaussians at levels 1 and 2 to a constant dark shade, simulating an underlying structure. Level 3 is brightened moderately, while levels 4 and 5 are brightened further, enhancing high-frequency details to create a glowing, skeletal appearance.

For the \textbf{Sharp} effect, we entirely remove level 2 and set the opacity of levels 3–5 to full, emphasizing fine details. To further enhance contrast, we darken the colors of Gaussians at level 5. This manipulation improves perceived sharpness while maintaining 3D consistency.

\end{document}